\documentclass[dvipsnames,table,10pt,twocolumn,letterpaper]{article}
 
\usepackage[pagenumbers]{iccv} 

\definecolor{iccvblue}{rgb}{0.21,0.49,0.74}
\usepackage[pagebackref,breaklinks,colorlinks,allcolors=iccvblue]{hyperref}

\usepackage{multirow}
\usepackage{subcaption} 
\usepackage{algorithmic}
\usepackage{amsmath}
\usepackage{algorithm}
\usepackage{tabularx}
\usepackage{xcolor}
\usepackage{mdframed}
\usepackage{stfloats}

\def\paperID{} 
\def\confName{ICCV}
\def\confYear{2025}

\title{Robust Machine Unlearning for Quantized Neural Networks via  Adaptive Gradient Reweighting with Similar Labels}

\author{
\begin{minipage}[t]{0.24\textwidth}
    \centering
    Yujia Tong \\
    Wuhan University of Technology \\
    Hubei Key Laboratory of Transportation Internet of Things\\
    {\tt\small tyjjjj@whut.edu.cn}
\end{minipage}
\hfill
\begin{minipage}[t]{0.24\textwidth}
    \centering
    Yuze Wang \\
    Wuhan University of Technology \\
    Hubei Key Laboratory of Transportation Internet of Things\\
    {\tt\small hbtmwyz@whut.edu.cn}
\end{minipage}
\hfill
\begin{minipage}[t]{0.24\textwidth}
    \centering
    Jingling Yuan\footnotemark[1] \\  
    Wuhan University of Technology \\
    Hubei Key Laboratory of Transportation Internet of Things\\
    {\tt\small yjl@whut.edu.cn}
\end{minipage}
\hfill
\begin{minipage}[t]{0.24\textwidth}
    \centering
    Chuang Hu \\
    State Key Laboratory of Internet of Things for Smart City \\
    University of Macau \\
    {\tt\small chuanghu@um.edu.mo}
\end{minipage}
}

\begin{document}
\maketitle
\footnotetext[1]{\ Corresponding author} 
\begin{abstract}
Model quantization enables efficient deployment of deep neural networks on edge devices through low-bit parameter representation, yet raises critical challenges for implementing machine unlearning (MU) under data privacy regulations. Existing MU methods designed for full-precision models fail to address two fundamental limitations in quantized networks: 1) Noise amplification from label mismatch during data processing, and 2) Gradient imbalance between forgotten and retained data during training. These issues are exacerbated by quantized models' constrained parameter space and discrete optimization. We propose Q-MUL, the first dedicated unlearning framework for quantized models. Our method introduces two key innovations: 1) Similar Labels assignment replaces random labels with semantically consistent alternatives to minimize noise injection, and 2) Adaptive Gradient Reweighting dynamically aligns parameter update contributions from forgotten and retained data. Through systematic analysis of quantized model vulnerabilities, we establish theoretical foundations for these mechanisms. Extensive evaluations on benchmark datasets demonstrate Q-MUL's superiority over existing approaches. 
\end{abstract}    
\section{Introduction}
\label{sec:intro}

Deep neural networks (DNNs) have achieved significant success in fields such as computer vision \cite{he2016deep} and natural language processing \cite{vaswani2017attention}, but their high demand for computation and memory limits their deployment on resource-constrained edge devices. Model quantization \cite{zhou2018adaptive,nagel2020up,fang2020post} reduces the storage requirements and computational complexity by lowering the bit-width of model weights and activations, making them more suitable for edge devices. To enhance performance, user-provided data can be utilized for periodic online fine-tuning of quantized models on edge devices. However, with the implementation of data privacy regulations such as GDPR \cite{hoofnagle2019european}, users have the right to request the deletion of their uploaded data, which involves not only removing data from databases but also ensuring that specific samples no longer influence the trained models. This raises a new challenge: \textbf{\textit{How to eliminate the influence of specific data samples on quantized models?}}

\begin{figure}[t]
    \centering
    \begin{subfigure}[b]{0.23\textwidth}
        \includegraphics[width=\textwidth]{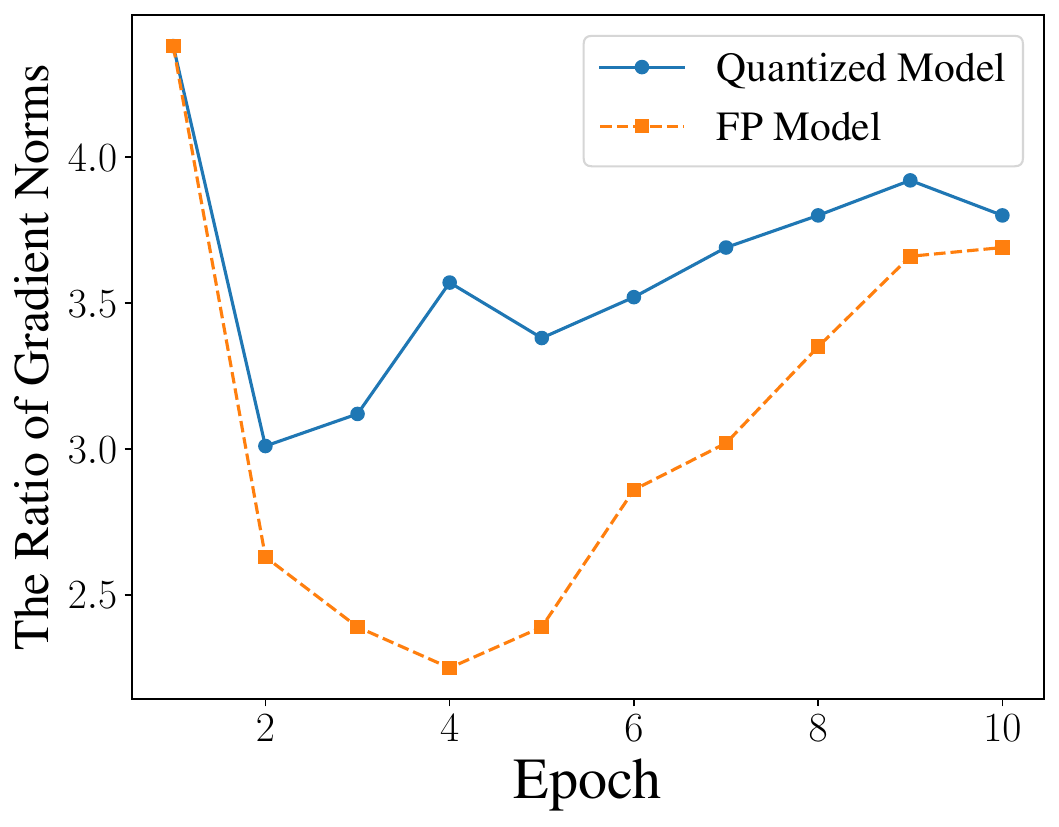}
        \caption{Comparison of gradient.}
        \label{fig:subfig1}
    \end{subfigure}
    \begin{subfigure}[b]{0.23\textwidth}
        \includegraphics[width=\textwidth]{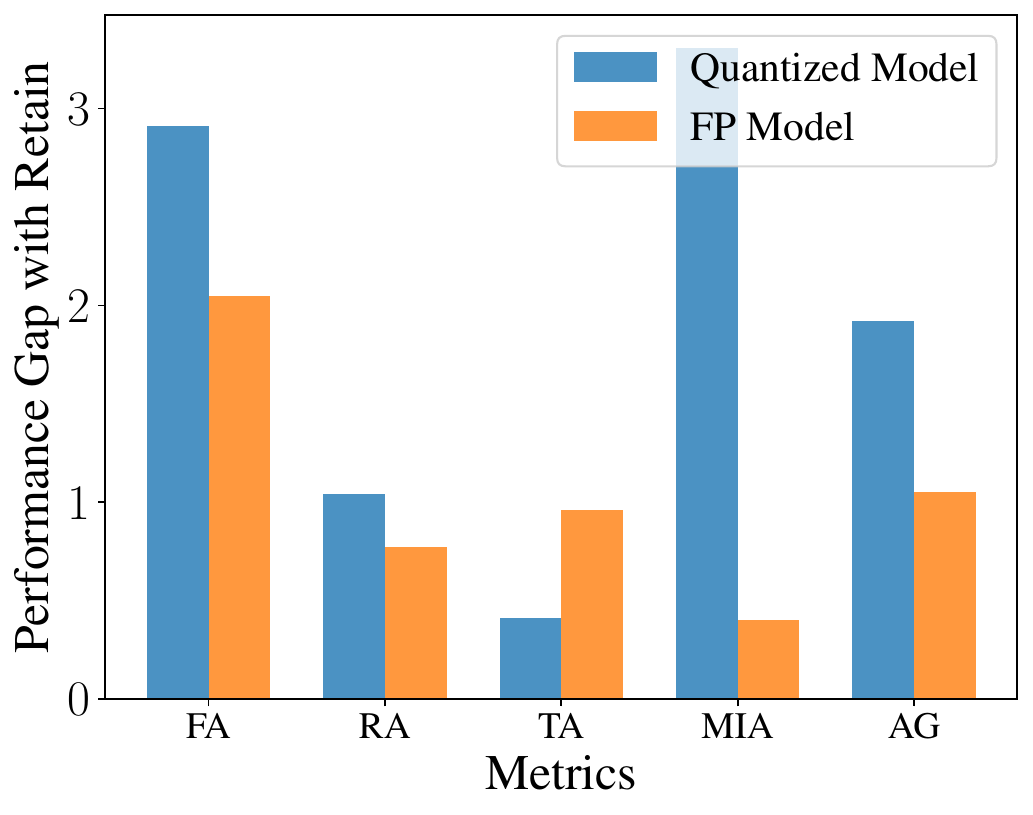}
        \caption{Comparison of performance.}
        \label{fig:subfig2}
    \end{subfigure}
    \caption{We conduct Random Labels\cite{golatkar2020eternal} on the full-precision and quantized models. (a) The ratio of gradient norms between the forgotten data and the retained data in the loss. (b) Performance gaps of FA, RA, TA, and MIA with Retrain, AG is the average gap. The observations are from ResNet18 on CIFAR-10.}
    \label{fig1}
     \vspace{-0.1mm}
\end{figure}

Machine unlearning (MU) provides a potential pathway to implement the \textit{right to be forgotten} through parametric influence removal. Existing MU approaches fall into two categories: 1) \textit{Exact unlearning} \cite{bourtoule2021machine} achieves ideal performance by retraining the model from scratch after removing the data to be forgotten, but it requires a high computational cost. 2) \textit{Approximate unlearning} \cite{chien2022certified} achieves comparable unlearning performance by adjusting the parameters of a pre-trained model, requiring fewer computational resources. \textit{We focus on approximate unlearning due to its speed and efficiency making it particularly well-suited for deployment on edge devices.}

Existing MU methods face two main limitations when applied to quantized models since they~\cite{golatkar2020eternal, fan2023salun} are primarily designed for full-precision models and do not adequately address the impact of quantization. Consequently, when these methods (e.g., Random Labels \cite{golatkar2020eternal} and SalUn \cite{fan2023salun}) are directly applied to quantized models, their effectiveness is reduced, particularly in two key stages: data processing and unlearning training. During the data processing phase, existing methods \cite{golatkar2020eternal,  fan2023salun} simulate the forgetting process by assigning random labels to samples in the forgotten dataset. However, the introduction of random labels introduces substantial noise that is inconsistent with the original data. Using these noisy samples in subsequent unlearning training can cause significant deviations in the direction of model parameter updates, thereby impairing the model's robustness. This issue is exacerbated in quantized models, as their limited parameter representation space makes them more sensitive to noise than full-precision models.

Furthermore, in the unlearning training phase, we observed a substantial discrepancy in the gradient norms when training with forgotten versus retained data. This discrepancy results in uneven contributions from the two datasets to the model’s parameter updates, thereby reducing the effectiveness of unlearning. While this issue is present in full-precision models, it is more pronounced in quantized models. As shown in Figure \ref{fig1}, the y-axis represents the ratio of gradient norms when using forgotten data to the gradient norms when using retained data. After unlearning, the average performance gap of quantized models compared to Retrain is 1.92\%, which is significantly higher than the 1.05\% gap observed in full-precision models (a smaller gap indicates a more effective forgetting process). This suggests that current methods lead to a degradation in the effectiveness of unlearning when applied to quantized models.

To address these challenges, we propose \textbf{Q-MUL}—a quantization-aware unlearning framework with dual innovations. During the data processing phase, we select the label closest to the true label as the label (that we term Similar Labels) for the samples in the forgotten dataset. During the unlearning training process, we adaptively weight the loss of the forgotten data and the retained data to balance their contributions to the updates of model parameters. We have conducted extensive experiments to validate the effectiveness of Q-MUL, and when applied to quantized models, its performance is superior to existing methods. We summarize our contributions below:

\begin{itemize}[leftmargin=*] \item To the best of our knowledge, we are the first to apply unlearning techniques directly to quantized models. We analyze the challenges involved, including the noise introduced during the data processing phase and the excessive gradient discrepancy between forgotten and retained data during unlearning training. 
\item We propose Q-MUL, a method that uses Similar Labels to replace the labels of forgotten data samples during the data processing phase, thereby reducing noise. In the unlearning training phase, we adaptively adjust the loss weights for forgotten and retained data based on the gradients, ensuring a balanced contribution to the model parameter updates. 
\item We conduct extensive experiments on benchmark datasets to evaluate the performance of Q-MUL. The results demonstrate that Q-MUL outperforms existing MU methods, making it more suitable for quantized models. \end{itemize}

\section{Related Work}
Our work bridges two distinct research domains: \textit{machine unlearning} for privacy compliance and \textit{model quantization} for efficient deployment. We analyze their intersection and identify critical gaps in existing literature.

\noindent\textbf{Machine Unlearning} has evolved from theoretical frameworks to practical implementations. The seminal work by Ginart et al. \cite{ginart2019making} formalized unlearning requirements for k-means clustering, while Bourtoule et al. \cite{bourtoule2021machine} introduced the "SISA" framework for deep neural networks through sharded training. Recent advances focus on approximate unlearning paradigms: Sekhari et al. \cite{sekhari2021remember} established theoretical guarantees for gradient-based updates, and Golatkar et al. \cite{golatkar2020eternal} proposed noise injection in parameter space for influence removal. However, these methods assume continuous parameter spaces (32-bit FP) and unrestricted gradient dynamics – assumptions invalidated by quantized models' discrete optimization landscapes. Notably, existing MU techniques exhibit three limitations in quantized contexts: 1) Random label assignment \cite{fan2023salun} introduces catastrophic noise in low-bit regimes, 2) Uniform gradient treatment \cite{golatkar2020eternal} fails to address the amplified imbalance caused by quantized representations, and 3) No existing work considers the interaction between backpropagation through quantizers and machine unlearning.

\noindent\textbf{Model Quantization} \cite{zhang2021diversifying,cai2020zeroq,xu2020generative}  is a key method in deep learning model compression technology, which reduces the size and computational requirements of the model by lowering the bit width of model parameters. Model quantization research has progressed along two main paths: Post-Training Quantization (PTQ) \cite{nagel2020up} that adjusts pre-trained models, and Quantization-Aware Training (QAT) \cite{esser2020learned} that embeds quantization during learning. PTQ directly quantizes the model after training without additional training, but it can lead to a significant degradation in model performance. QAT simulates quantization operations during the model training process, allowing the model to adapt to the errors introduced by quantization from the outset, which can alleviate the decline in model performance. PACT~\cite{choi2018pact} employs a learnable clipping threshold optimized during training for low-precision quantization. DSQ \cite{gong2019differentiable} introduces a differentiable soft tanh function to approximate standard quantization steps. As the most widely used and representative QAT method, LSQ+ \cite{bhalgat2020lsq+} uses learnable quantization parameters to achieve better accuracy. Since MU methods typically require training with data, unlearning training can be viewed as a process of Quantization-Aware Training. This paper is committed to proposing an unlearning approach suitable for Quantization-Aware Training.

\section{Preliminaries}
In this section, we first revisit the concept of \textit{machine unlearning}. Subsequently, we introduce the fundamental computational process of \textit{Quantization-Aware Training}.

\subsection{Revisiting Machine Unlearning}
Let the complete (training) dataset \( {D}=\{ (x_i,y_i) \}_{i=1}^{N}\)  be with N number of samples, in which \( x_i \in \mathbb{R}^d \) is the \( {i}^{th} \) sample with the associated class label \( y_i \in \{1,2,...,K\}\). The forgotten dataset \( {D}_f \subseteq {D} \) represents a subset of the complete dataset \( D \), typically denoting the data that requires deletion from the trained model. And we denote its complement as retained dataset \( D_r \), which represents the data we want to retain, \( i.e. \), \( D_f \cap D_r = \varnothing \), \( D_f \cup D_r = {D}\). Based on the composition of \(D_f\), machine unlearning (MU) in image classification can be divided into two categories:  class-wise forgetting and random data forgetting. In class-wise forgetting, the data in \( D_f\) consists entirely of samples from the same class, with the purpose of eliminating the influence of an entire class on the model. In random data forgetting, the data in \(D_f\) may include random samples from one or multiple classes, with the purpose of eliminating the influence of randomly selected data points from \( D_f\)  on the model. Prior to unlearning, we denote the \( \mathcal{M}_0 \) as the original model. In the context of MU, we denote retrained model \( \mathcal{M}_{r} \) as the model trained from scratch solely on the retained dataset, which is regarded as the ``golden standard"~\cite{nguyen2022survey}. Nonetheless, retraining entails a substantial computational overhead. Therefore, approximate unlearning was proposed, with the primary goal of  obtaining unlearned model \( \mathcal{M}_{u} \) by eliminating the influence of forgotten dataset \( D_f \) from the original model \( \mathcal{M}_{0} \), thereby approximating retrained model \( \mathcal{M}_{r} \) with less computational overhead. 

\subsection{Quantization-Aware Training}
Quantization-Aware Training (QAT) employs fake quantization nodes to simulate quantization error during inference. These nodes perform quantization and dequantization operations on floating-point numbers.  For \( n \)-bit signed quantization,  given the scale factor \( s \), we can define a quantization function \( q(\cdot) \) to represent the operations within the fake quantization node:

\begin{equation}
\label{eq1}
x^q = q(x^r) = s \times \left\lfloor \text{clamp}\left(\frac{x^r}{s}, -2^{n-1}, 2^{n-1} - 1 \right) \right\rceil,
\end{equation}
where \( \left\lfloor \cdot \right\rceil \) rounds to the nearest integer, and \( \text{clamp}(x, r_{\text{low}}, r_{\text{high}}) \) ensures \( x \) is within \([r_{\text{low}}, r_{\text{high}}]\). In neural networks with quantized activations and weights, forward and backward propagation can be summarized as:
\begin{align}
\label{eq2}
    &\text{Forward:} \quad \text{Output}(x) = x^q \cdot w^q = q(x^r) \cdot q(w^r),\notag\\
    &\text{Backward:} \quad \frac{\partial \mathcal{L}}{\partial x^r} = \left\{
    \begin{array}{ll}
    \frac{\partial \mathcal{L}}{\partial x^q} & \text{if } x \in [-Q_N^x, Q_P^x] \\
    0 & \text{otherwise}
    \end{array}
    \right., \\
    &\quad\quad\quad\quad\quad \frac{\partial \mathcal{L}}{\partial {\bf w^r}} = \left\{
    \begin{array}{ll}
    \frac{\partial \mathcal{L}}{\partial {\bf w^q}} & \text{if } {\bf w} \in [-Q_N^{\bf w}, Q_P^{\bf w}] \\
    0 & \text{otherwise}\notag
    \end{array}
    \right.,
\end{align}
where \( \mathcal{L} \) is the loss function, \( q(\cdot) \) is used during forward propagation, and the straight-through estimator (STE) \cite{bengio2013estimating} is employed for gradient derivation during backward propagation. In image classification tasks, neural networks map input images \( x \) to predicted probability distributions \( p(x; w) \) via the forward function \( f(x; w) \). For any sample \( (x_i,y_i) \) in the dataset \( {D}=\{ (x_i,y_i) \}_{i=1}^{N}\), the quantized model's output probability distribution (logits) is:
\begin{equation}
\label{eq3}
  p_{\mathbf{Q}}(x_i^q;w^q)=f(x_i^q;w^q),
\end{equation}

In order to simplify the notation, we will use \( x_i \) to denote the input and \( w \) to denote the quantized model weights in the following text.

\section{Methodology}

In this section, we provide a comprehensive introduction to the proposed Q-MUL. We first introduce the overview of Q-MUL. Subsequently, we provide a detailed description of two components of Q-MUL: \textit{Similar Labels} and \textit{Adaptive Gradient Reweighting}.

\subsection{The Overview of Q-MUL}
\begin{figure*}[t]
\centering
\includegraphics[width=0.96\linewidth]{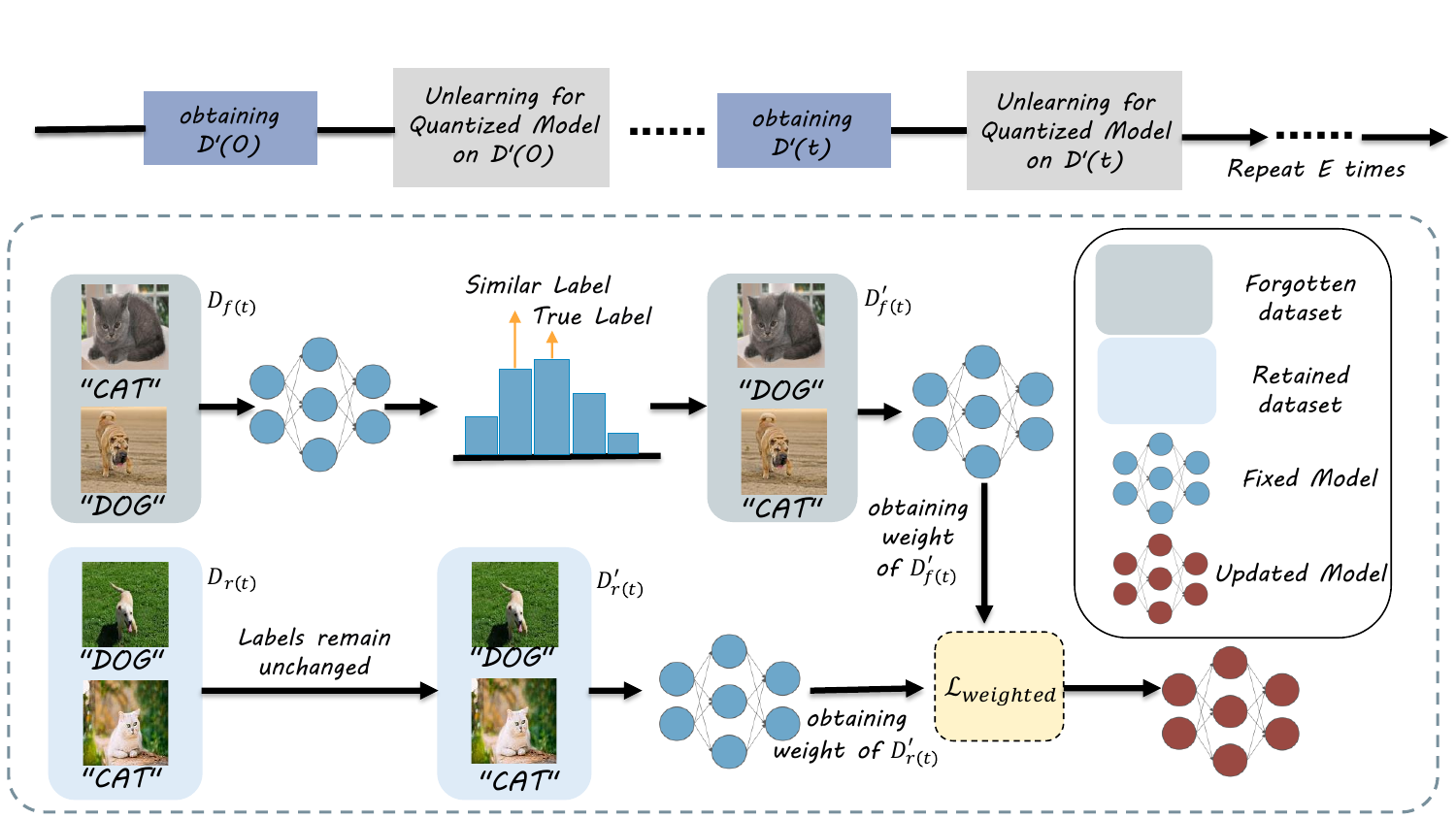}
\caption{The overview of Q-MUL. Before performing the unlearning process on the model, Q-MUL first selects the label that is closest to the true label in the current output of the model as the new label for the data samples that need to be forgotten. During the unlearning process, we adaptively calculate the weights of the forgotten and retained data based on their gradient norms.}
\label{fig2}
\end{figure*}

In Figure \ref{fig2}, we present an overview of Q-MUL, which is divided into two stages: data processing and unlearning training. During the data processing stage, Q-MUL selects the label closest to the true label( \textbf{Similar Labels}) as the training label for the samples in the forgotten dataset $D_f$, while keeping the labels of the samples in the retained dataset $D_r$ unchanged, resulting in the processed datasets $D'_f$ and $D'_r$ for the next stage of training. In the unlearning training stage, for each epoch, Q-MUL first calculates the gradient norms of the loss function on datasets $D'_f$ and $D'_r$, and then adaptively calculates their weights based on the gradient norms and incorporates them into the loss function. After several epochs of training, an unlearned model is obtained. In the next two subsections of this section, we will provide a detailed introduction to the two stages of Q-MUL.

\subsection{Similar Labels}
In MU, to achieve the forgetting of specific samples (samples in the forgotten dataset), it is common to train by randomly shuffling the labels of the samples in the forgotten dataset\cite{golatkar2020eternal,fan2023salun}. However, this method has some obvious shortcomings. Random labels (RL) do not take into account the potential relationships between data, introducing a large amount of noise inconsistent with the original data. Assume that the true label of the data to be forgotten is \textit{“cat”}. If we use a random label, such as \textit{“car”}, which is clearly unrelated to the original data, this may lead to a significant difference in the direction of parameter updates when the model is trained on these data compared to using the original label, thereby affecting the overall stability and Robustness of the model, causing it to forget information that should not be forgotten.

To address these issues, we propose Similar Labels (SL). SL selects labels that are closest to the original label, thereby achieving the forgetting of specific data while reducing the negative impact on model performance. This method helps to maintain the consistency and logic of the data because it reduces the amount of noise introduced and can better maintain the model's predictive ability for similar samples. Moreover, SL can more precisely control the forgetting process because it aligns the direction of model parameter updates (gradient direction) with the true labels, thereby avoiding the introduction of substantial random noise. This helps to mitigate the potential over-adjustment of the model during the unlearning process.

Specifically, we first input the forgetting data into the model. For a forgetting instance \((x_i, y_i) \in D_f\) , the model's predicted probability distribution is \( p_{\mathbf{Q}}( x_i;w) \). The probability predicted by the model corresponding to the true label is \( p_{\text{gt}} = p_{\mathbf{Q}}(y_i \mid x_i; w ) \). For each class \(k \in \{1,2,..., K\}\)of the model output, we calculate the distance between its predicted probability and the true label probability \( d(k) \):
\begin{equation}
\label{eq4}
d(k) = \left| p_{\mathbf{Q}}( k \mid {x}_i; w ) - p_{\text{gt}} \right|,
\end{equation}
Then, we find the similar label \( k_{sl} \) that minimizes the distance \( d(k) \):
\begin{equation}
\label{eq5}
k_{sl} = \underset{k \neq y_i}{\arg\min} \; d(k).
\end{equation}
Finally, we use these similar labels \( k_{sl} \) instead of the original labels \( y_i \). To better understand the selection of Similar Labels, we provide the pseudocode of the selection process in Algorithm \ref{alg1}.

We conduct a theoretical analysis by comparing the differences in gradient update directions between SL and RL. Let the model parameters be \( w \), the original label of sample \( x_i \) be \( y_i \), the similar label be \( k_{sl} \), and the random label be \( k_{rl} \). During the training process, the cross-entropy loss can be expressed as: 
\begin{equation}
\mathcal{L} = -\sum_{(x_i, y_i) \in D} \sum_{k=1}^{K} y_{i,k} \log p_Q(k \mid x_i; w)
\end{equation}
where \( y_{i,k} \) is the one-hot encoding of the true label \( y_i \), i.e., \( y_{i,k} = 1 \) if and only if \( k = y_i \), otherwise \( y_{i,k} = 0 \). Therefore, the loss function can be simplified as follows:
\begin{equation}
\mathcal{L} = -\sum_{(x_i, y_i) \in D} \log p_\mathbf{Q}(y_i \mid x_i; w)
\end{equation}
where \( p_\mathbf{Q}(y_i \mid x_i; w) \) is the probability that the model predicts the true class \( y_i \) for the sample \( x_i \). For the sample  \((x_i,  y_i) \), the gradient of the loss function is:
\begin{equation}
\nabla_w \mathcal{L} = -\nabla_w \log p_{\mathbf{Q}}(y_i \mid x_i; w)
\end{equation}
We measure the difference in gradient directions using cosine similarity. The cosine of the angle \( \theta_{sl} \) between the gradients when using SL and the original label is:
\begin{equation}
\small
\cos \theta_{sl} = \frac{\nabla_w \log p_{\mathbf{Q}}(y_i \mid x_i;w) \cdot \nabla_w \log p_{\mathbf{Q}}(k_{sl} \mid x_i;w)}{\|\nabla_w \log p_{\mathbf{Q}}(y_i \mid x_i;w)\| \cdot \|\nabla_w \log p_{\mathbf{Q}}(k_{sl} \mid x_i;w)\|}
\end{equation}
The cosine of the angle \( \theta_{rl} \) between the gradients when using RL and the original label has an approximate form. Since the selection of \( k_{sl} \) makes \( p_{\mathbf{Q}}(k_{sl} \mid x_i; w) \) close to \( p_{\text{gt}} \), the angle \( \theta_{sl} \) between the gradient directions of SL and the true label is small, hence \( \cos \theta_{sl} \approx 1 \). This indicates that the gradient update direction of similar labels is highly consistent with the gradient direction of the true label, avoiding the gradient conflicts that may arise from random labels. From the perspective of gradient update directions, this ensures the robustness of quantized models after unlearning.

\begin{algorithm}[t]
\small
\caption{Similar Labels}
\label{alg1}
\begin{algorithmic}[1]
    \REQUIRE Original complete dataset \( {D}=\{ (x_i,y_i) \}_{i=1}^{N} \), The quantized model \( \mathcal{M}_{t} \) with parameters \({w}_{t}\)at epoch t
    \ENSURE Training dataset \( D' \) for unlearning
   \STATE initialize  \({D}_{f}^{'}={\emptyset} \),  \({D}_{r}^{'}={\emptyset} \);
   \FOR{\( i \in  [1, ..., N] \)}
        \IF{\( (x_i,y_i) \in {D}_f \)}
            \STATE Calculate predicted probability  \( p_{\mathbf{Q}}(x_i;w)  \) by eq.(\ref{eq3});
            \STATE Calculate distance \( d(k) \) following eq.(\ref{eq4});
            \STATE Calculate class label \( k_{sl} \) by eq.(\ref{eq5}) ;
            \STATE Add  \( (x_i,k_{sl}) \) to   \( {D}_{f}^{'} \);
        \ELSE
            \STATE Add  \( (x_i,y_i) \) to   \( {D}_{r}^{'} \);
        \ENDIF
    \ENDFOR
    \STATE \(D'\)=\({D}_{r}^{'} \cup {D}_{f}^{'}\);
\RETURN \( D' \)
\end{algorithmic}
\end{algorithm}

\subsection{Adaptive Gradient Reweighting}
In methods based on RL~\cite{golatkar2020eternal, fan2023salun}, the model is trained using the forgotten data with labels randomly shuffled and the retained data with labels unchanged. However, these methods are designed for full-precision models, and their performance significantly degrades when applied to quantized models. Our experiments reveal that the degradation is due to the large gradient differences between the forgotten and retained data during the unlearning process, leading to inconsistent contributions to model parameter updates. As shown in Figure \ref{fig1}, we apply the RL method to both full-precision and 4-bit quantized models for unlearning and compared the gradient norms of model parameters when trained with forgetting and retained data in each epoch (the y-axis represents the gradient norm when using forgotten data divided by the gradient norm when using retained data). Due to the limited parameter representation range, the differences in gradient norms when training with forgotten and retained data are larger in quantized models than in full-precision models, leading to inconsistent contributions to model parameter updates. Therefore, the performance of these methods degrades when applied to quantized models.

To address this issue, we propose Adaptive Gradient Reweighting (AGR). AGR adaptively adjusts the contributions of forgotten data and retained data to model updates to achieve balance. Specifically, we use a weighted cross-entropy loss function to adjust sample weights:
\begin{equation}
\label{eq10}
\mathcal{L}_{\text{weighted}}(x, y;w) = \frac{1}{|D'|} \sum_{(x_i,y_i)\in D'} \alpha_i \cdot \mathcal{L}_{\text{CE}}(x_i, y_i;w),
\end{equation}
where \(\mathcal{L}_{\text{CE}}\) is the cross-entropy loss function used to calculate the loss of each sample, \(D'\) is the training dataset for unlearning, \(y\) is the true label and \(\alpha_{i}\) are the weights of each sample used to adjust the contribution of each sample to the total loss. To achieve adaptive adjustment, we design a gradient norm balancing strategy. Before each training epoch, the expected gradient norms of the two types of data are calculated separately:
\begin{align}
\label{eq11}
G_f &= \mathbb{E}_{(x,y) \sim {D'}_{\text{f}}} \left[ \| \nabla_{w} \mathcal{L}_{\text{forget}}(x, y;w) \|_2 \right] \\
\label{eq12}
G_r &= \mathbb{E}_{(x,y) \sim {D'}_{\text{r}}} \left[ \| \nabla_{w} \mathcal{L}_{\text{retain}}(x,y;w) \|_2 \right]
\end{align}
where  \(\mathcal{L}_{\text{forget}}\) and \(\mathcal{L}_{\text{retain}}\) are the loss functions for the forgotten and retained data, respectively. The weight coefficients are dynamically allocated based on the gradient norms:
\begin{equation}
\label{eq13}
\alpha_f = \frac{G_r}{G_f + G_r}, \quad \alpha_r = \frac{G_f}{G_f + G_r}
\end{equation}
During the forgetting training process, when using Stochastic Gradient Descent (SGD) to update the model parameters, the parameter update at time $t$ can be expressed as:
\begin{equation}
\label{eq14}
\mathbf{w}_{t+1} = \mathbf{w}_t - \eta_t (\alpha_f \nabla{\mathcal{L}_{\text{forget}}} + \alpha_r \nabla{\mathcal{L}_{\text{retain}}})
\end{equation}
In SGD, model parameters are updated based on the gradient direction, while the gradient magnitude (norm) determines the update step size. If the gradient magnitudes of different data points vary significantly, the model may overemphasize certain data points while neglecting others, leading to instability. 
Therefore, the weighted gradient $\nabla{\mathcal{L}_{\text{weighted}}}$ is obtained by combining the gradients of the forgotten and retained data with their respective weights. If the gradient norm of the forgotten data is greater than that of the retained data, then the weight for the forgotten data will be relatively smaller, and vice versa. This complementarity ensures a more balanced contribution of both types of data in the model parameter updates, thereby enhancing the robustness of the quantized model after unlearning.

To help better understand the unlearning process, we provide the overall pipeline of Q-MUL in Algorithm \ref{alg2}. 

\begin{algorithm}[tb]
\small
\caption{The overall pipeline of Q-MUL}
\label{alg2}
\begin{algorithmic}[1] 
\REQUIRE The original quantized model ${\mathcal{M}}_{o}$ with parameters \({w}_0\) , Original complete dataset \( {D}=\{ (x_i,y_i) \}_{i=1}^{N} \), Total training epochs $T$ 
\ENSURE The unlearned quantized model ${\mathcal{M}}_{u}$ with parameters \({w}_u\)   
\FOR{$t \in [0, \ldots, T-1]$}
    \STATE Process the original dataset \( D \) to obtain the training dataset \( D' \) following Algorithm \ref{alg1};
    \STATE Calculate gradient norms \(G_f\)  by eq.(\ref{eq11});
    \STATE Calculate gradient norms \(G_r\) by eq.(\ref{eq12});
    \STATE Calculate weight of \(  {D'}_r \) and \( {D'}_f  \)  by eq.(\ref{eq13}); 
    \STATE Update the parameters \( w_t\) of quantized model ${\mathcal{M}}_{t}$  by eq.(\ref{eq14});
\ENDFOR 
\RETURN Unlearned Quantized model ${\mathcal{M}}_{u}$ with parameters \({w}_u\) 
\end{algorithmic}
\end{algorithm}
\section{Experiments}
\subsection{Experimental Setup}
\noindent\textbf{Datasets and Networks.} We choose CIFAR-10 \cite{krizhevsky2009learning}, CIFAR-100 \cite{krizhevsky2009learning}, SVHN \cite{netzer2011reading} and Tiny-Imagenet\cite{le2015tiny} as the datasets. We evaluate our method on two mainstream deep neural networks, MobileNetV2 \cite{howard2017mobilenets} and ResNet-18 \cite{he2016deep}. We quantize the weights of MobileNetV2 to 2 bits, while keeping the activations in full-precision, and quantize both the weights and activations of ResNet-18 to 4 bits.

\noindent\textbf{Baselines.}
To demonstrate the superiority of Q-MUL, we choose different types of baselines, which can be classified as: \textbf{Retrain}: Using the retained dataset to retrain a model from scratch is the most effective method for machine unlearning, yet it consumes a significant amount of time. Fine-Tuning (\textbf{FT}) \cite{warnecke2021machine,golatkar2020eternal}:  Using the retained dataset to slightly fine-tune the original model, unlike retraining, FT requires only a few training epochs. Gradient Ascent (\textbf{GA}) \cite{graves2021amnesiac,thudi2022unrolling}: Contrary to gradient descent, GA trains on the forget dataset and updates the parameters of the original model in the direction of gradient ascent. Influence Unlearning (\textbf{IU}) \cite{koh2017understanding,izzo2021approximate}: Using influence functions to estimate the impact of \( D_f \) on the \( \mathcal{M}_0 \), and then applying a Newton-one-step-update to the parameters, eliminating the influence of specific data points. Random Labels (\textbf{RL}) \cite{golatkar2020eternal}: Train on the complete dataset after assigning random labels to the targets in the forgotten dataset. \textbf{$\ell_1$-sparse} \cite{liu2024model}: Infuse weight sparsity into approximate unlearning through model pruning. \textbf{SalUn} \cite{fan2023salun}: Integrate the gradient-based weight saliency map with the RL method. 

\begin{table*}[htbp]
  \centering
  \small
\setlength{\tabcolsep}{0.6mm}{
    \begin{tabular}{c|ccccc|ccccc}
\toprule    
\multirow{2}[4]{*}{Method}  & \multicolumn{5}{c|}{CIFAR-10} & \multicolumn{5}{c}{CIFAR-100} \\
\cmidrule{2-11}
 & FA & RA & TA & MIA & AG$\color{blue}\downarrow$  & FA & RA & TA & MIA & AG$\color{blue}\downarrow$ \\
    \midrule
     \multicolumn{11}{c}{\cellcolor{gray!20}The proportion of forgotten data samples to all samples is 10\%} \\
    \midrule
    Retrain  & 93.38 & 100.0 & 92.64 & 13.36 &\cellcolor{gray!20} 0 & 74.76 & 99.98 & 72.43 & 56.36 &\cellcolor{gray!20} 0 \\
    FT  & 99.42 \textcolor{blue}{(6.04)} & 99.96 \textcolor{blue}{(0.04)} & 93.05 \textcolor{blue}{(0.41)} & 2.82 \textcolor{blue}{(10.54)} &\cellcolor{gray!20} 4.26  & 91.82 \textcolor{blue}{(17.06)} & 94.12 \textcolor{blue}{(5.86)} & 65.29 \textcolor{blue}{(7.14)} & 16.69 \textcolor{blue}{(39.67)} & \cellcolor{gray!20} 17.43\\
   GA & 99.38 \textcolor{blue}{(6.00)} & 99.38 \textcolor{blue}{(0.62)} & 93.35 \textcolor{blue}{(0.71)} & 1.22 \textcolor{blue}{(12.14)} &\cellcolor{gray!20} 4.87    & 95.02 \textcolor{blue}{(20.26)} & 96.86 \textcolor{blue}{(3.12)} & 71.84 \textcolor{blue}{(0.59)} & 9.56 \textcolor{blue}{(46.80)} &\cellcolor{gray!20} 17.69\\
    IU & 95.64 \textcolor{blue}{(2.26)}& 95.53\textcolor{blue}{(4.47)} & 87.85\textcolor{blue}{(4.79)} & 7.93\textcolor{blue}{(5.43)} &\cellcolor{gray!20} 4.24 & 3.69\textcolor{blue}{(71.07)} & 4.03\textcolor{blue}{(95.95)} & 3.85\textcolor{blue}{(68.58)} & 98.8 \textcolor{blue}{(42.44)}&\cellcolor{gray!20} 69.51\\
    RL  &96.29  \textcolor{blue}{(2.91)}&98.96 \textcolor{blue}{(1.04)} &92.23  \textcolor{blue}{(0.41)} &16.67  \textcolor{blue}{(3.31)} &1.92\cellcolor{gray!20}  
    & 68.51  \textcolor{blue}{(6.25)}& 98.91 \textcolor{blue}{(1.07)} & 69.47\textcolor{blue}{(2.96)} & 85.62 \textcolor{blue}{(29.26)}& \cellcolor{gray!20} 9.89 \\
    $\ell_1$-sparse & 98.71 \textcolor{blue}{(5.33)} & 99.88 \textcolor{blue}{(0.12)} & 92.92 \textcolor{blue}{(0.28)} & 5.29 \textcolor{blue}{(8.07)} &\cellcolor{gray!20} 3.45 & 88.20 \textcolor{blue}{(13.44)} & 99.65 \textcolor{blue}{(0.33)} & 70.80 \textcolor{blue}{(1.63)} & 29.64 \textcolor{blue}{(26.72)} &\cellcolor{gray!20} 10.53 \\
    SalUn  & 96.82 \textcolor{blue}{(3.44)} & 99.45 \textcolor{blue}{(0.55)} & 92.32 \textcolor{blue}{(0.32)} & 14.27\textcolor{blue}{(0.91)} &\cellcolor{gray!20}  \underline{1.31}  & 82.22 \textcolor{blue}{(7.46)}& 98.71 \textcolor{blue}{(1.27)}& 67.38 \textcolor{blue}{(5.05)}&  66.78\textcolor{blue}{(10.42)} &\cellcolor{gray!20}\underline{  6.05 }\\
    Q-MUL & 97.11\textcolor{blue}{(3.73)} & 99.67\textcolor{blue}{(0.33)} & 92.39\textcolor{blue}{(0.25)} & 12.49\textcolor{blue}{(0.87)} &\cellcolor{gray!20} \textbf{ 1.30 } &75.71\textcolor{blue}{(0.95)}  &97.89\textcolor{blue}{(2.09)}  &67.27\textcolor{blue}{(5.16)}  &52.11\textcolor{blue}{(4.25)}  &\textbf{3.11}\cellcolor{gray!20}  \\
    \midrule
    \multicolumn{11}{c}
    {\cellcolor{gray!20}The proportion of forgotten data samples to all samples is 30\%} \\
    \midrule
    Retrain  & 91.61 & 99.97 & 91.29 & 16.11 &\cellcolor{gray!20} 0  & 67.67 & 99.98 & 67.54 & 58.47 &\cellcolor{gray!20} 0  \\
    FT  & 99.19 \textcolor{blue}{(7.58)} & 99.95 \textcolor{blue}{(0.02)} & 92.94 \textcolor{blue}{(1.65)} & 3.10 \textcolor{blue}{(13.01)} &\cellcolor{gray!20} 5.57 & 96.24 \textcolor{blue}{(28.57)} & 99.94 \textcolor{blue}{(0.04)} & 71.49 \textcolor{blue}{(3.95)} & 20.57 \textcolor{blue}{(37.90)} &\cellcolor{gray!20} 17.62  \\
    GA & 99.35 \textcolor{blue}{(7.74)} & 99.38 \textcolor{blue}{(0.59)} & 93.39 \textcolor{blue}{(2.10)} & 1.28 \textcolor{blue}{(14.83)} &\cellcolor{gray!20} 6.32 & 96.73 \textcolor{blue}{(29.06)} & 97.30 \textcolor{blue}{(2.68)} & 73.35 \textcolor{blue}{(5.81)} & 6.19 \textcolor{blue}{(52.28)} &\cellcolor{gray!20} 22.46\\
    IU & 76.21 \textcolor{blue}{(15.40)}& 75.93\textcolor{blue}{(24.04)} & 71.48\textcolor{blue}{(19.81)}  & 24.53\textcolor{blue}{(8.42)} &\cellcolor{gray!20} 16.92 &2.81\textcolor{blue}{(64.86)} & 3.19\textcolor{blue}{(96.79)} & 3.09\textcolor{blue}{(64.45)} & 96.33\textcolor{blue}{(37.86)}  &\cellcolor{gray!20} 65.99\\
    RL  &  93.96 \textcolor{blue}{(2.35)} & 96.03\textcolor{blue}{(3.94)} & 90.34 \textcolor{blue}{(0.95)}& 19.07 \textcolor{blue}{(2.96)} &\cellcolor{gray!20}
     2.55  & 75.50 \textcolor{blue}{(7.83)}& 99.58\textcolor{blue}{(0.40)} & 66.56\textcolor{blue}{(0.98)} & 87.33 \textcolor{blue}{(28.86)}&\cellcolor{gray!20}  9.52\\
    $\ell_1$-sparse & 99.06 \textcolor{blue}{(7.45)} & 99.90 \textcolor{blue}{(0.07)} & 93.01 \textcolor{blue}{(1.72)} & 4.54 \textcolor{blue}{(11.57)} &\cellcolor{gray!20} 5.20 & 88.59 \textcolor{blue}{(20.92)} & 99.79 \textcolor{blue}{(0.19)} & 70.72 \textcolor{blue}{(3.18)} & 33.16 \textcolor{blue}{(25.31)} &\cellcolor{gray!20} 12.40 \\
    SalUn  & 96.17\textcolor{blue}{(4.56)} & 97.78\textcolor{blue}{(2.19)} & 91.57\textcolor{blue}{(0.30)} & 15.50\textcolor{blue}{(0.66)} &\cellcolor{gray!20} \underline{1.93} & 79.60 \textcolor{blue}{(11.93)} & 96.70 \textcolor{blue}{(3.28)} & 63.41 \textcolor{blue}{(4.13)} & 57.73 \textcolor{blue}{(0.74)} &\cellcolor{gray!20}\underline{  5.02 }\\
    Q-MUL &93.57 \textcolor{blue}{(1.96)} &96.77 \textcolor{blue}{(3.20)} &90.24 \textcolor{blue}{(1.05)} &15.59 \textcolor{blue}{(0.52)} &\textbf{1.68}\cellcolor{gray!20} &73.90\textcolor{blue}{(6.23)}  &97.63\textcolor{blue}{(2.35)}  &65.80\textcolor{blue}{(1.74)}  &63.76\textcolor{blue}{(5.29)}  &\textbf{3.90}\cellcolor{gray!20}   \\
    \midrule
    \multicolumn{11}{c}{\cellcolor{gray!20}The proportion of forgotten data samples to all samples is 50\%} \\
    \midrule
    Retrain  & 90.75 & 100.0 & 90.28 & 18.76 &\cellcolor{gray!20} 0   & 65.40 & 99.99 & 65.74 & 68.19 &\cellcolor{gray!20} 0  \\
    FT  & 99.29 \textcolor{blue}{(8.54)} & 99.96 \textcolor{blue}{(0.04)} & 93.28 \textcolor{blue}{(3.00)} & 2.85 \textcolor{blue}{(15.91)} &\cellcolor{gray!20} 6.87  & 97.10 \textcolor{blue}{(31.70)} & 99.98 \textcolor{blue}{(0.01)} & 72.82 \textcolor{blue}{(7.08)} & 16.86 \textcolor{blue}{(51.33)} &\cellcolor{gray!20} 22.71\\
   GA  & 99.34 \textcolor{blue}{(8.59)} & 99.34 \textcolor{blue}{(0.66)} & 93.44 \textcolor{blue}{(3.16)} & 1.27 \textcolor{blue}{(17.49)} &\cellcolor{gray!20} 7.48  & 96.82 \textcolor{blue}{(31.42)} & 97.33 \textcolor{blue}{(2.66)} & 73.37 \textcolor{blue}{(7.63)} & 5.93 \textcolor{blue}{(62.26)} &\cellcolor{gray!20} 25.99\\
    IU  & 16.12\textcolor{blue}{(74.63)} & 16.02\textcolor{blue}{(83.98)} & 16.22\textcolor{blue}{(74.06)} & 16.36\textcolor{blue}{(2.40)} &\cellcolor{gray!20} 58.77 & 0.96\textcolor{blue}{(64.44)} & 1.04\textcolor{blue}{(98.95)} & 1.00\textcolor{blue}{(64.74)} & 98.96\textcolor{blue}{(30.77)} &\cellcolor{gray!20} 64.73\\
    RL  & 95.32 \textcolor{blue}{(4.57)}& 97.17 \textcolor{blue}{(2.83)}& 90.85 \textcolor{blue}{(0.57)}& 21.33 \textcolor{blue}{(2.57)}&\cellcolor{gray!20} 2.64 & 69.20 \textcolor{blue}{(3.80)} & 85.60 \textcolor{blue}{(14.39)}& 64.39 \textcolor{blue}{(1.35)}& 54.58 \textcolor{blue}{(13.61)} &\cellcolor{gray!20}  8.29 \\
    $\ell_1$-sparse & 99.25 \textcolor{blue}{(8.50)} & 99.95 \textcolor{blue}{(0.05)} & 92.67 \textcolor{blue}{(2.39)} & 3.91 \textcolor{blue}{(14.85)} &\cellcolor{gray!20} 6.45  & 96.48 \textcolor{blue}{(31.08)} & 99.98 \textcolor{blue}{(0.01)} & 72.26 \textcolor{blue}{(6.52)} & 25.11 \textcolor{blue}{(43.08)} & 20.17\cellcolor{gray!20}   \\
    SalUn  & 96.56 \textcolor{blue}{(5.81)} & 98.03 \textcolor{blue}{(1.97)} & 91.55 \textcolor{blue}{(1.27)} &18.05 \textcolor{blue}{(0.71)} &\cellcolor{gray!20}
     \underline{2.44} & 87.75 \textcolor{blue}{(22.35)} & 98.99 \textcolor{blue}{(1.00)} & 64.69\textcolor{blue}{(1.05)}   & 73.12\textcolor{blue}{(4.93)}  &\cellcolor{gray!20} \underline{  7.33 }\\
    Q-MUL  &91.09  \textcolor{blue}{(0.34)} &93.94 \textcolor{blue}{(6.06)} &89.12 \textcolor{blue}{(1.16)} &18.68   
    \textcolor{blue}{(0.08)} &\cellcolor{gray!20}\textbf{ 1.91} &67.25\textcolor{blue}{(1.85)}  &89.65\textcolor{blue}{(10.34)}  &61.38\textcolor{blue}{(4.36)}  &56.52\textcolor{blue}{(11.67)}  & \textbf{ 7.06}\cellcolor{gray!20}  \\
    \bottomrule
    \end{tabular} %
    }
    \caption{Performance of various MU methods for MobileNetV2 with activations kept at full precision and weights quantized to 2 bits on
CIFAR-10 and CIFAR-100. The unlearning scenario is random data forgetting. \textbf{Bold} indicates the best performance and \underline{underline} indicates the runner-up. A performance gap against Retrain is provided in \textcolor{blue}{(•)}.  }
  \label{tab1}%
\end{table*}

\begin{table}[htbp]
  \centering
  \small
  \setlength{\tabcolsep}{0.12mm}{
    \begin{tabular}{c|ccccc}
\toprule   
Method 
  & FA & RA & TA & MIA & AG$\color{blue}\downarrow$  \\
    \midrule
    Retrain &74.76		&99.98		&72.43		&56.36    &\cellcolor{gray!20} 0 \\
    Q-MUL  &75.71 \textcolor{blue}{(0.95)} &97.89 \textcolor{blue}{(2.09)} & 67.27\textcolor{blue}{(5.16)} &52.11 \textcolor{blue}{(4.25)}  & 3.11\cellcolor{gray!20}  \\
    \textit{\textbf{w/o}}SL &73.67 \textcolor{blue}{(1.09)} &96.46 \textcolor{blue}{(3.52)} &65.91 \textcolor{blue}{(6.52)} & 51.09\textcolor{blue}{(5.27)} &4.10\cellcolor{gray!20}  \\
    \textit{\textbf{w/o}}AGR &69.84 \textcolor{blue}{(4.92)} &92.08 \textcolor{blue}{(7.90)} &68.62 \textcolor{blue}{(3.81)} &58.00 \textcolor{blue}{(1.64)} &4.57\cellcolor{gray!20}  \\
    \bottomrule
    \end{tabular}%
    }
    \caption{ Impact of Removing
 Components of Q-MUL. Ablation study is from ResNet18 on CIFAR-100. The unlearning scenario is random data forgetting(10\%). }
  \label{tab2}
\end{table}

\noindent\textbf{Evaluation Metrics.} Consistent with $\ell_1$-sparse \cite{liu2024model} and SalUn \cite{fan2023salun}, we select multiple evaluation metrics, as follows: Forget Accuracy (\textbf{FA}), which evaluates the model's accuracy on the forget dataset after undergoing machine unlearning. Retain Accuracy (\textbf{RA}), which evaluates the model's accuracy on the retained dataset after undergoing machine unlearning. Test Accuracy (\textbf{TA}), which measures the model's accuracy on the testing dataset, which can be used to assess the generalization ability of the model after undergoing MU.
Membership Inference Attack (\textbf{MIA}) ~\cite{shokri2017membership}, which is an attack method aimed at determining whether a particular data sample was used to train a given machine learning model. MIA can be used to assess whether a model still retains residual information of certain data after attempting to forget them. Please note that the above four metrics are not simply the higher the better or the lower the better, but rather, they should have as little gap as possible compared to the \textbf{Retrain}. Finally, we introduce the Average Gap (\textbf{AG}), which calculates the average gaps between the baseline method and Retrain in terms of the four metrics: FA, RA, TA, and MIA after unlearning. The closer the AG is to 0, the better the unlearning effect.

\noindent\textbf{Implementation details.} We use the LSQ+ \cite{bhalgat2020lsq+} quantization method unless specified. The original model is trained from scratch for 182 epochs using SGD with a cosine-annealed learning rate (initial 0.1). The Retrain method follows the same optimizer configuration but reduces training to 160 epochs. We conduct experiments under two unlearning scenarios: random data forgetting and class-wise forgetting, with a primary focus on random data forgetting due to its prevalence in real-world applications. Other MU training details and additional experimental results are provided in the Appendix.

\subsection{Performance Evaluation}
As presented in Table \ref{tab1}, we demonstrate the performance of various MU methods for quantized ResNet18 under random data forgetting scenario. Compared to all the baseline methods, the performance of Q-MUL is the best. Specifically, On CIFAR-10, existing methods have already achieved relatively small average gaps, but Q-MUL is still able to further reduce the gap compared to Retrain. On CIFAR-100, when the proportion of forgotten data samples to all samples is 10\%, the FA and RA of Q-MUL are 75.71\% and 97.89\% respectively (with gaps of 0.95\% and 2.09\% from Retrain), while the previous state-of-the-art (SOTA) method SalUn has FA and RA of 82.22\% and 98.71\% respectively (with gaps of 7.46\% and 1.27\% from Retrain). This indicates that Q-MUL balances the contributions of forgotten and retained data to the model parameter updates. For the MIA metric, Q-MUL is 52.11\% (with a gap of 4.25\% from Retrain) and SalUn is 66.78\% (with a gap of 10.42\% from Retrain), showing that the forgetting effect of Q-MUL is closer to the optimal machine unlearning method Retrain. Although the TA metric of Q-MUL has a slightly higher gap than SalUn, the average gap of the four metrics from Retrain is 3.11\%, which is significantly lower than the existing baseline methods. When the proportion of forgotten data samples to all samples is 30\% and 50\%, Q-MUL can also achieve the smallest average gaps of 3.90\% and 7.06\%.

In the appendix, we provide experiments on quantized ResNet18 on the larger dataset Tiny-Imagenet and the class-wise forgetting scenario, as well as experiments on quantized MobileNetV2 across different datasets.

\begin{figure*}[t]
    \centering
    \begin{subfigure}[b]{0.19\textwidth}
        \includegraphics[width=\textwidth]{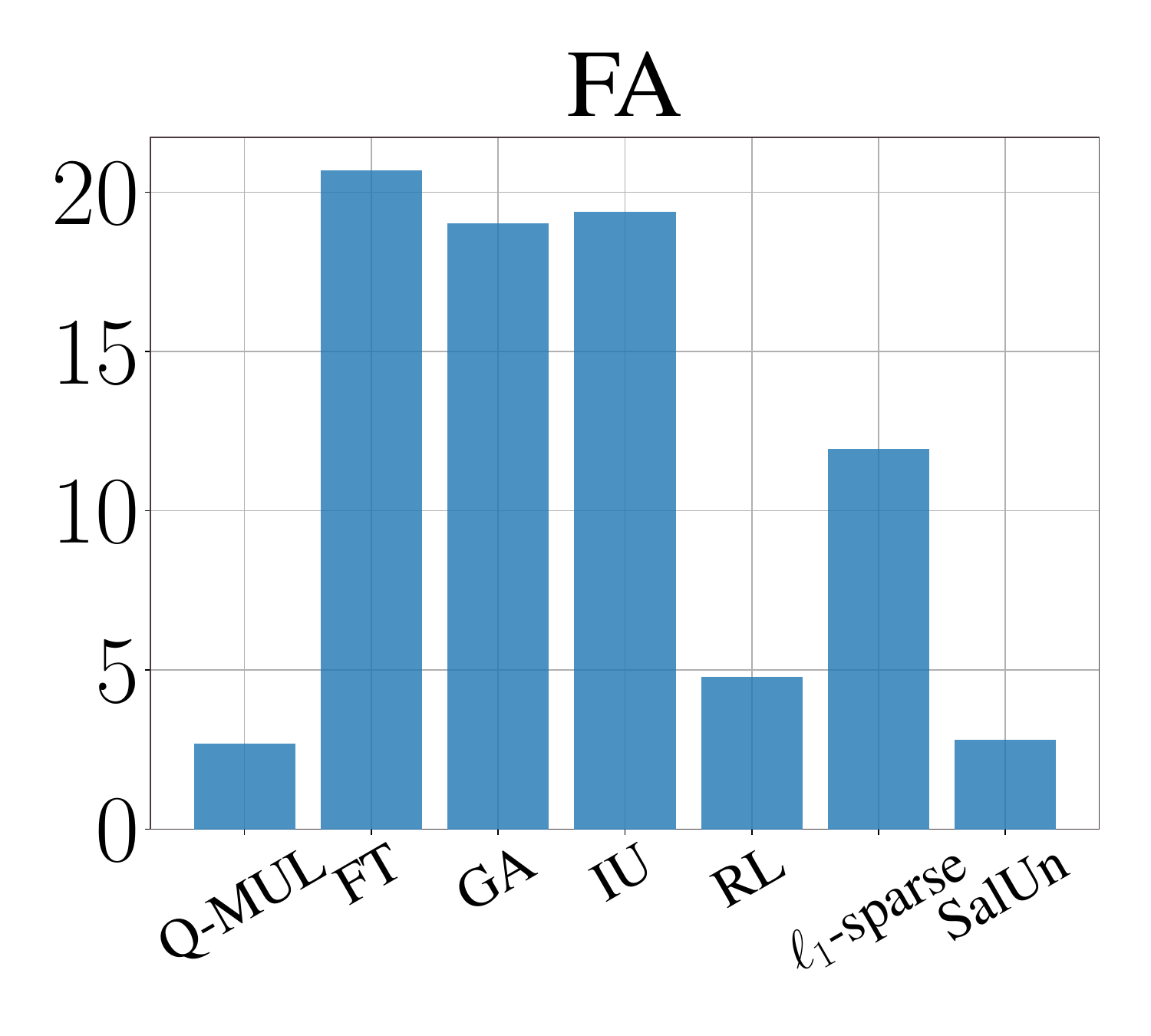}
        
        \label{fig:subfig111}
    \end{subfigure}
    \begin{subfigure}[b]{0.19\textwidth}
        \includegraphics[width=\textwidth]{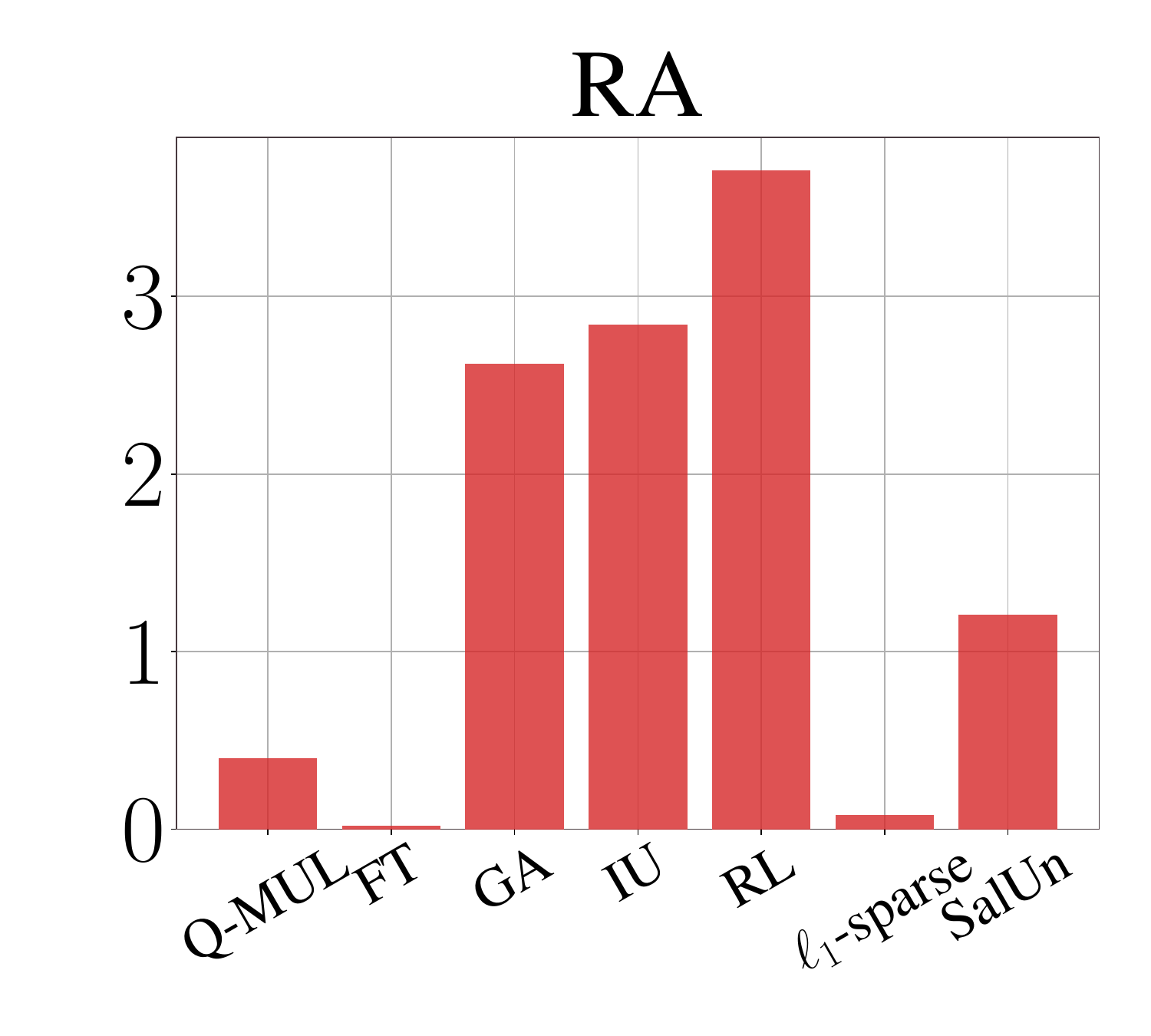}
        
        \label{fig:subfig222}
    \end{subfigure}
    \begin{subfigure}[b]{0.19\textwidth}
        \includegraphics[width=\textwidth]{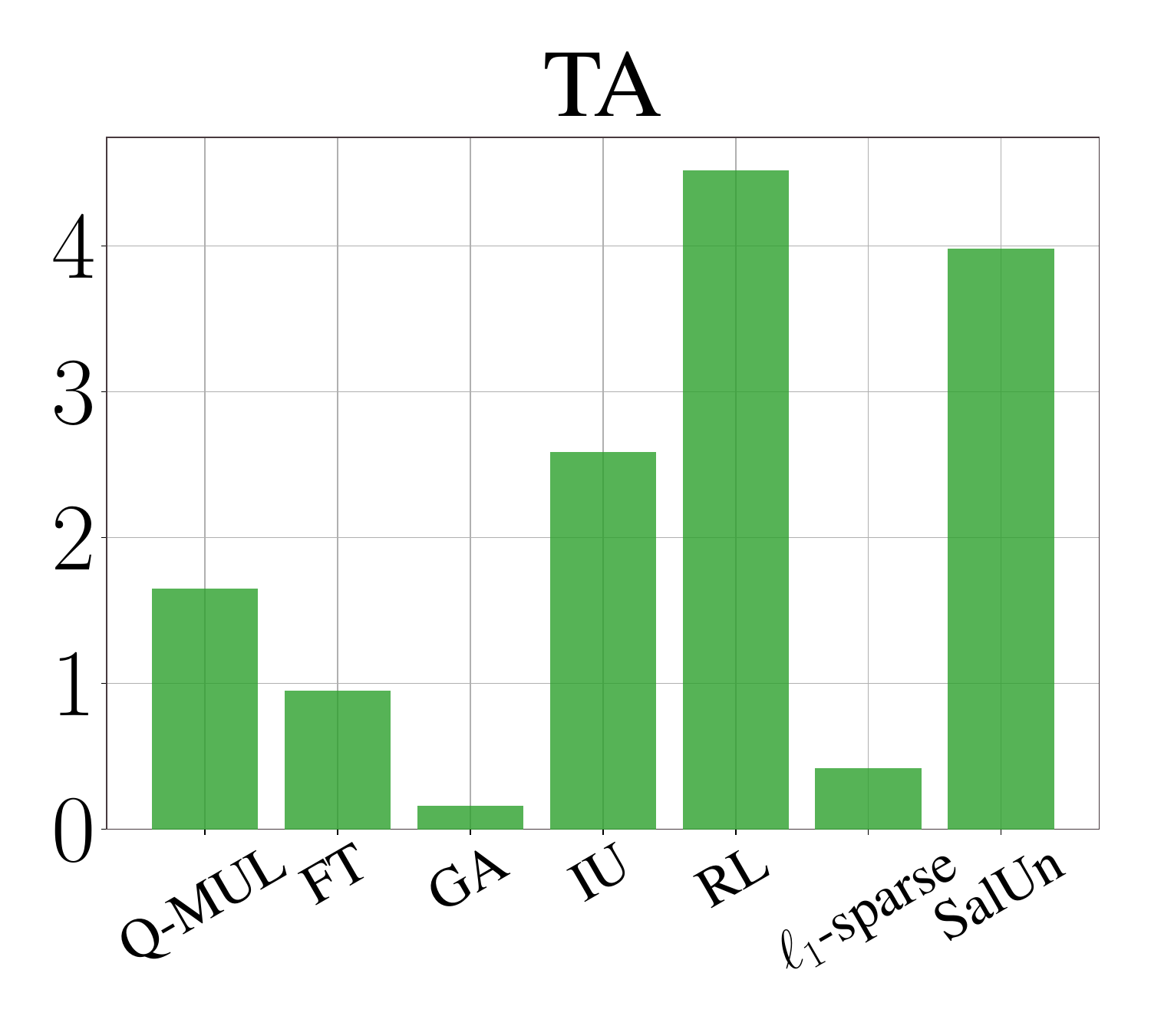}
        
        \label{fig:subfig3}
    \end{subfigure}
    \begin{subfigure}[b]{0.19\textwidth}
        \includegraphics[width=\textwidth]{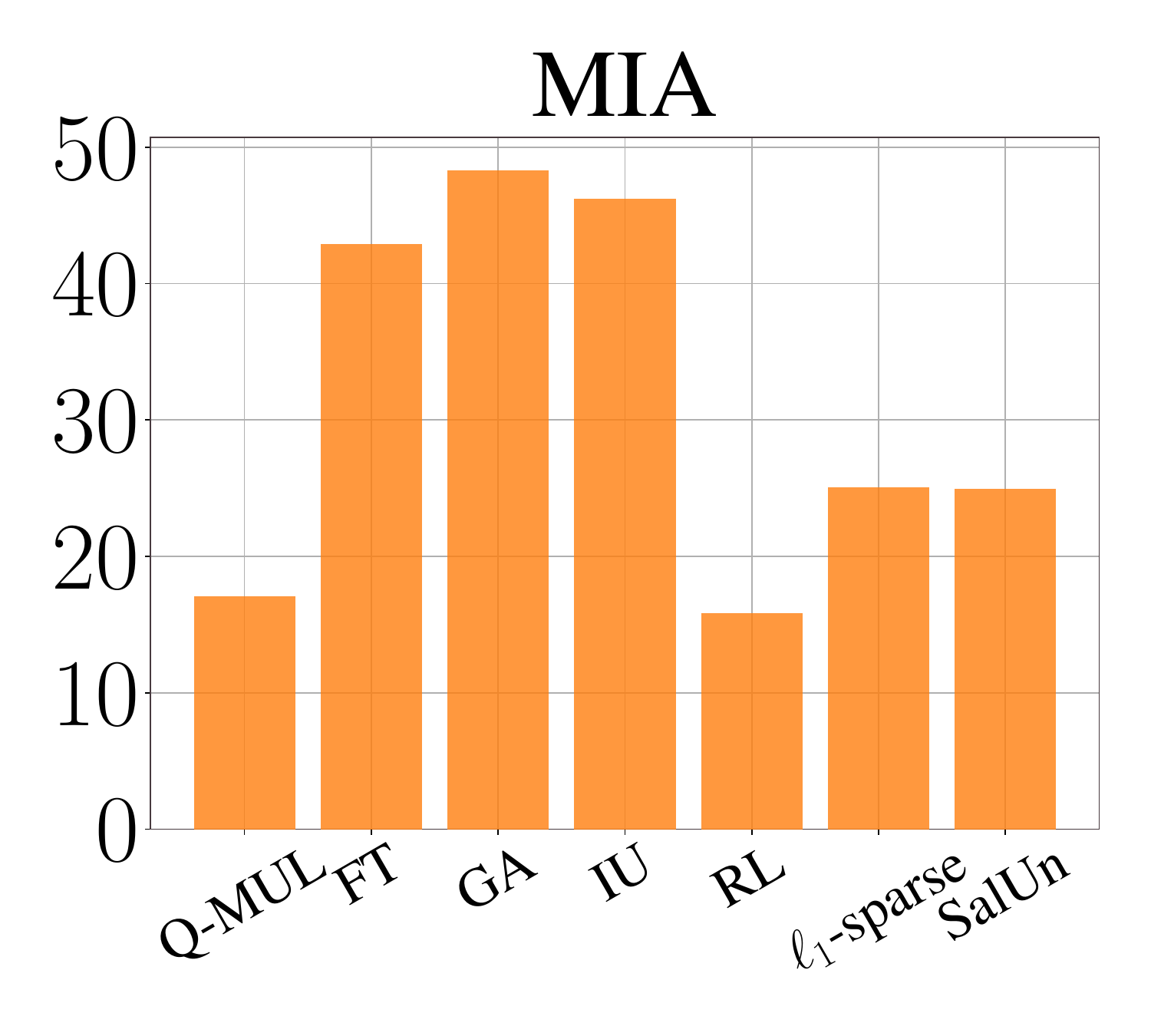}
        
        \label{fig:subfig4}
    \end{subfigure}
    \begin{subfigure}[b]{0.19\textwidth}
        \includegraphics[width=\textwidth]{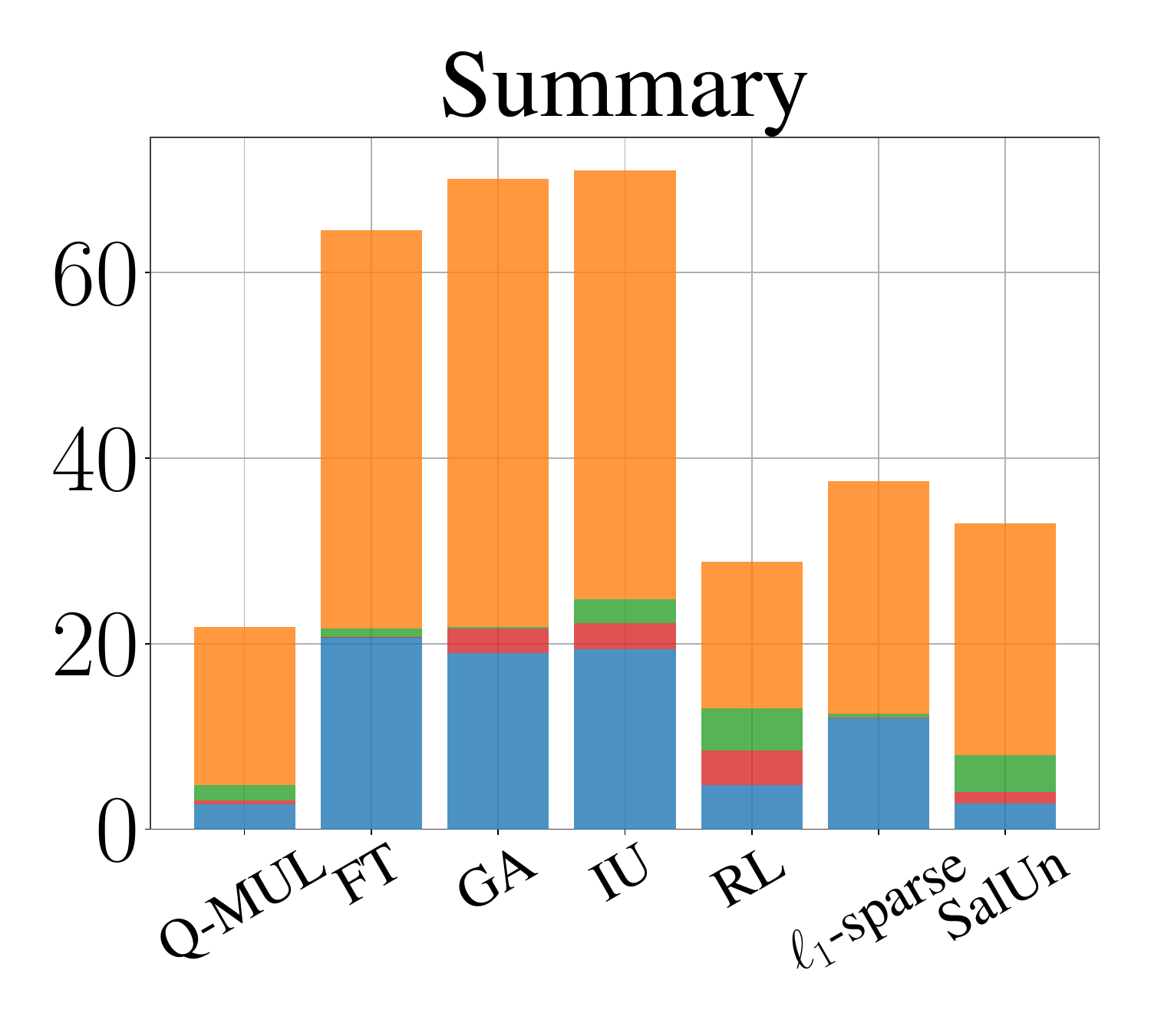}
        
        \label{fig:subfig15}
    \end{subfigure}
    \caption{Performance gaps of Full-precision ResNet18 on CIFAR-100. The unlearning scenario is random data forgetting (10\%). A shorter bar (smaller gap) indicates performance is closer to that of the Retrained model. The average gap for each method is calculated by dividing the values in the Summary bar chart by 4. }
    \label{fig3}
\end{figure*}

\begin{table*}[htbp]
  \centering
  \small
    \begin{tabular}{c|c|c|c|ccccc}
\toprule QAT &${D}_f/{D}$ &Bit Width   
&MU 
  & FA & RA & TA & MIA & AG$\color{blue}\downarrow$  \\
  \midrule
  \multirow{5}{*}{PACT\cite{choi2018pact}} & \multirow{5}{*}{10\%} &4w4a &Retrain &66.64 &94.90 &65.84 &40.78 &0\cellcolor{gray!20} \\
        &&4w4a & RL  &71.87\textcolor{blue}{(5.23)}  &80.28\textcolor{blue}{(14.62)}  &60.56\textcolor{blue}{(5.28)}  &34.18\textcolor{blue}{(6.60)}  &\underline{7.93}\cellcolor{gray!20}  \\
     &&4w4a &$\ell_1$-sparse  &78.69\textcolor{blue}{(12.05)}  &85.42\textcolor{blue}{(9.48)}  &65.90\textcolor{blue}{(0.06)}  &24.44\textcolor{blue}{(16.34)}  &9.48\cellcolor{gray!20}  \\
     &&4w4a &SalUn &75.13\textcolor{blue}{(8.49)}  &81.21\textcolor{blue}{(13.69)}  &61.16\textcolor{blue}{(4.68)}  &30.60\textcolor{blue}{(10.18)}  &9.26\cellcolor{gray!20}\\
     &&4w4a &Q-MUL &73.51\textcolor{blue}{(6.87)}  &84.73\textcolor{blue}{(10.17)}  &66.10\textcolor{blue}{(0.26)}  & 34.09\textcolor{blue}{(6.69)} &\textbf{6.00}\cellcolor{gray!20}\\
    \midrule
    \multirow{5}{*}{DSQ \cite{gong2019differentiable}} & \multirow{5}{*}{10\%} &4w4a &Retrain &65.24 &99.22 &  63.62 & 44.42& 0\cellcolor{gray!20} \\
    &&4w4a 
     & RL  &80.42\textcolor{blue}{(15.18)}   & 90.28\textcolor{blue}{(8.94)} &60.54\textcolor{blue}{(3.08)}  &32.09\textcolor{blue}{(12.33)}  & \underline{9.88}\cellcolor{gray!20}  \\
     &&4w4a 
     &$\ell_1$-sparse  & 89.11\textcolor{blue}{(23.87)} & 97.11\textcolor{blue}{(2.11)} & 65.36\textcolor{blue}{(1.74)} & 19.82\textcolor{blue}{(24.60)} & 13.08\cellcolor{gray!20}  \\
     &&4w4a &SalUn & 82.73\textcolor{blue}{(17.49)} & 87.04\textcolor{blue}{(12.18)} & 58.50\textcolor{blue}{(5.12)} & 21.75\textcolor{blue}{(22.67)} & 14.37 \cellcolor{gray!20}\\
     &&4w4a &Q-MUL &78.18\textcolor{blue}{(12.94)}  &92.28\textcolor{blue}{(6.94)} &65.31\textcolor{blue}{(1.69)} & 32.89\textcolor{blue}{(11.53)} &\cellcolor{gray!20}\textbf{8.26}\\
        
    \bottomrule 
    \end{tabular}%
     \caption{ ResNet18 with both activations and weights quantized to 4 bit on CIFAR-100. We conduct unlearning across different QAT methods. The unlearning scenario is random data forgetting(10\%). }
  \label{tab3}
\end{table*}

\subsection{Ablation Studies}
\textbf{The Impact of Different Components. } To further demonstrate the effectiveness of different  components of Q-MUL, we conduct ablation experiments using two variants of Q-MUL: (a) \textit{\textbf{w/o}}  Similar Labels (SL)  and (b) \textit{\textbf{w/o}}  Adaptive Gradient Reweighting (AGR). As shown in Table \ref{tab2}, using the complete Q-MUL enables the performance of the unlearned model to be closest to that of the retrained model, with an average gap of only 3.11\%. When SL is removed, the gap in the RA  metric increases from 2.09\% to 3.52\%, indicating a decline in the discriminative ability of the unlearned model on the retained data. This is attributed to the introduction of noise inconsistent with the original data during the data processing phase, which leads to excessive model adjustments and the forgetting of information that should not have been forgotten. When AGR is removed, the forgotten model's FA and RA are 69.84\% and 92.08\%, respectively, with gaps of 4.92\% and 7.90\% compared to the retrained model's FA and RA (which increased by 3.97\% and 5.81\% when using AGR, respectively). This indicates that AGR is crucial for balancing the contributions of forgotten and retained data to the model parameter updates.

\noindent \textbf{Q-MUL with Full-precision Models.} To more comprehensively evaluate Q-MUL, we compare the effects of different unlearning methods on the full-precision model, as shown in Figure  \ref{fig3}. The issues of noise introduction and gradient imbalance also exist in the full-precision model, although they are not as severe as in the quantized model. However, Q-MUL can still address these two issues. Specifically, the average gap of Q-MUL can reach 5.46\%, which is significantly lower than that of existing methods. This demonstrates that Q-MUL can achieve superior unlearning performance even in full-precision models.

\noindent \textbf{Q-MUL with Different QAT Methods.} In addition to LSQ+, we also validate the performance of different MU methods under other QAT methods. As shown in Table \ref{tab3}, Q-MUL can achieve SOTA performance in other QAT methods as well. Specifically, for PACT, the gaps between Q-MUL and Retrain in FA  and TA are 6.87\% and 10.17\%, respectively, with an average gap of 6.00\%. For DSQ, the gaps between Q-MUL and Retrain in FA and TA are 12.94\% and 6.94\%, respectively, with an average gap of 8.26\%. Q-MUL effectively reduces the RA gap while balancing the FA gap and RA gap as much as possible.

\section{Conclusions}
This paper is the first to directly apply machine unlearning to quantized models. We analyze the challenges of machine unlearning in quantized models: noise introduced during the data processing phase and the large gradient difference of the loss function between forgotten and retained data. To address these challenges, we propose Q-MUL. This method reduces noise by replacing the labels of forgotten data with similar labels during the data processing phase. During training, it adaptively adjusts the weights of the loss from forgotten and retained data based on the gradient to balance their contributions to model parameter updates. Experimental results show that Q-MUL is more suitable for quantized models than existing methods.

{
    \small
    \bibliographystyle{ieeenat_fullname}
    \bibliography{main}

\begin{thebibliography}{32}
\providecommand{\natexlab}[1]{#1}
\providecommand{\url}[1]{\texttt{#1}}
\expandafter\ifx\csname urlstyle\endcsname\relax
  \providecommand{\doi}[1]{doi: #1}\else
  \providecommand{\doi}{doi: \begingroup \urlstyle{rm}\Url}\fi

\bibitem[Bengio et~al.(2013)Bengio, L{\'e}onard, and Courville]{bengio2013estimating}
Yoshua Bengio, Nicholas L{\'e}onard, and Aaron Courville.
\newblock Estimating or propagating gradients through stochastic neurons for conditional computation.
\newblock \emph{arXiv preprint arXiv:1308.3432}, 2013.

\bibitem[Bhalgat et~al.(2020)Bhalgat, Lee, Nagel, Blankevoort, and Kwak]{bhalgat2020lsq+}
Yash Bhalgat, Jinwon Lee, Markus Nagel, Tijmen Blankevoort, and Nojun Kwak.
\newblock Lsq+: Improving low-bit quantization through learnable offsets and better initialization.
\newblock In \emph{Proceedings of the IEEE/CVF conference on computer vision and pattern recognition workshops}, pages 696--697, 2020.

\bibitem[Bourtoule et~al.(2021)Bourtoule, Chandrasekaran, Choquette-Choo, Jia, Travers, Zhang, Lie, and Papernot]{bourtoule2021machine}
Lucas Bourtoule, Varun Chandrasekaran, Christopher~A Choquette-Choo, Hengrui Jia, Adelin Travers, Baiwu Zhang, David Lie, and Nicolas Papernot.
\newblock Machine unlearning.
\newblock In \emph{2021 IEEE Symposium on Security and Privacy (SP)}, pages 141--159. IEEE, 2021.

\bibitem[Cai et~al.(2020)Cai, Yao, Dong, Gholami, Mahoney, and Keutzer]{cai2020zeroq}
Yaohui Cai, Zhewei Yao, Zhen Dong, Amir Gholami, Michael~W Mahoney, and Kurt Keutzer.
\newblock Zeroq: A novel zero shot quantization framework.
\newblock In \emph{Proceedings of the IEEE/CVF conference on computer vision and pattern recognition}, pages 13169--13178, 2020.

\bibitem[Chien et~al.(2022)Chien, Pan, and Milenkovic]{chien2022certified}
Eli Chien, Chao Pan, and Olgica Milenkovic.
\newblock Certified graph unlearning.
\newblock \emph{arXiv preprint arXiv:2206.09140}, 2022.

\bibitem[Choi et~al.(2018)Choi, Wang, Venkataramani, Chuang, Srinivasan, and Gopalakrishnan]{choi2018pact}
Jungwook Choi, Zhuo Wang, Swagath Venkataramani, Pierce I-Jen Chuang, Vijayalakshmi Srinivasan, and Kailash Gopalakrishnan.
\newblock Pact: Parameterized clipping activation for quantized neural networks.
\newblock \emph{arXiv preprint arXiv:1805.06085}, 2018.

\bibitem[Esser et~al.(2020)Esser, McKinstry, Bablani, Appuswamy, and Modha]{esser2020learned}
Steven~K Esser, Jeffrey~L McKinstry, Deepika Bablani, Rathinakumar Appuswamy, and Dharmendra~S Modha.
\newblock Learned step size quantization.
\newblock In \emph{International Conference on Learning Representations}, 2020.

\bibitem[Fan et~al.(2023)Fan, Liu, Zhang, Wei, Wong, and Liu]{fan2023salun}
Chongyu Fan, Jiancheng Liu, Yihua Zhang, Dennis Wei, Eric Wong, and Sijia Liu.
\newblock Salun: Empowering machine unlearning via gradient-based weight saliency in both image classification and generation.
\newblock \emph{arXiv preprint arXiv:2310.12508}, 2023.

\bibitem[Fang et~al.(2020)Fang, Shafiee, Abdel-Aziz, Thorsley, Georgiadis, and Hassoun]{fang2020post}
Jun Fang, Ali Shafiee, Hamzah Abdel-Aziz, David Thorsley, Georgios Georgiadis, and Joseph~H Hassoun.
\newblock Post-training piecewise linear quantization for deep neural networks.
\newblock In \emph{Computer Vision--ECCV 2020: 16th European Conference, Glasgow, UK, August 23--28, 2020, Proceedings, Part II 16}, pages 69--86. Springer, 2020.

\bibitem[Ginart et~al.(2019)Ginart, Guan, Valiant, and Zou]{ginart2019making}
Antonio Ginart, Melody Guan, Gregory Valiant, and James~Y Zou.
\newblock Making ai forget you: Data deletion in machine learning.
\newblock \emph{Advances in neural information processing systems}, 32, 2019.

\bibitem[Golatkar et~al.(2020)Golatkar, Achille, and Soatto]{golatkar2020eternal}
Aditya Golatkar, Alessandro Achille, and Stefano Soatto.
\newblock Eternal sunshine of the spotless net: Selective forgetting in deep networks.
\newblock In \emph{Proceedings of the IEEE/CVF Conference on Computer Vision and Pattern Recognition}, pages 9304--9312, 2020.

\bibitem[Gong et~al.(2019)Gong, Liu, Jiang, Li, Hu, Lin, Yu, and Yan]{gong2019differentiable}
Ruihao Gong, Xianglong Liu, Shenghu Jiang, Tianxiang Li, Peng Hu, Jiazhen Lin, Fengwei Yu, and Junjie Yan.
\newblock Differentiable soft quantization: Bridging full-precision and low-bit neural networks.
\newblock In \emph{Proceedings of the IEEE/CVF international conference on computer vision}, pages 4852--4861, 2019.

\bibitem[Graves et~al.(2021)Graves, Nagisetty, and Ganesh]{graves2021amnesiac}
Laura Graves, Vineel Nagisetty, and Vijay Ganesh.
\newblock Amnesiac machine learning.
\newblock In \emph{Proceedings of the AAAI Conference on Artificial Intelligence}, pages 11516--11524, 2021.

\bibitem[He et~al.(2016)He, Zhang, Ren, and Sun]{he2016deep}
Kaiming He, Xiangyu Zhang, Shaoqing Ren, and Jian Sun.
\newblock Deep residual learning for image recognition.
\newblock In \emph{Proceedings of the IEEE conference on computer vision and pattern recognition}, pages 770--778, 2016.

\bibitem[Hoofnagle et~al.(2019)Hoofnagle, Van Der~Sloot, and Borgesius]{hoofnagle2019european}
Chris~Jay Hoofnagle, Bart Van Der~Sloot, and Frederik~Zuiderveen Borgesius.
\newblock The european union general data protection regulation: what it is and what it means.
\newblock \emph{Information \& Communications Technology Law}, 28\penalty0 (1):\penalty0 65--98, 2019.

\bibitem[Howard et~al.(2017)Howard, Zhu, Chen, Kalenichenko, Wang, Weyand, Andreetto, and Adam]{howard2017mobilenets}
Andrew~G Howard, Menglong Zhu, Bo Chen, Dmitry Kalenichenko, Weijun Wang, Tobias Weyand, Marco Andreetto, and Hartwig Adam.
\newblock Mobilenets: Efficient convolutional neural networks for mobile vision applications.
\newblock \emph{arXiv preprint arXiv:1704.04861}, 2017.

\bibitem[Izzo et~al.(2021)Izzo, Smart, Chaudhuri, and Zou]{izzo2021approximate}
Zachary Izzo, Mary~Anne Smart, Kamalika Chaudhuri, and James Zou.
\newblock Approximate data deletion from machine learning models.
\newblock In \emph{International Conference on Artificial Intelligence and Statistics}, pages 2008--2016. PMLR, 2021.

\bibitem[Koh and Liang(2017)]{koh2017understanding}
Pang~Wei Koh and Percy Liang.
\newblock Understanding black-box predictions via influence functions.
\newblock In \emph{International conference on machine learning}, pages 1885--1894. PMLR, 2017.

\bibitem[Krizhevsky et~al.(2009)Krizhevsky, Hinton, et~al.]{krizhevsky2009learning}
Alex Krizhevsky, Geoffrey Hinton, et~al.
\newblock Learning multiple layers of features from tiny images.
\newblock 2009.

\bibitem[Le and Yang(2015)]{le2015tiny}
Yann Le and Xuan Yang.
\newblock Tiny imagenet visual recognition challenge.
\newblock \emph{CS 231N}, 7\penalty0 (7):\penalty0 3, 2015.

\bibitem[Liu et~al.(2024)Liu, Ram, Yao, Liu, Liu, SHARMA, Liu, et~al.]{liu2024model}
Jiancheng Liu, Parikshit Ram, Yuguang Yao, Gaowen Liu, Yang Liu, PRANAY SHARMA, Sijia Liu, et~al.
\newblock Model sparsity can simplify machine unlearning.
\newblock \emph{Advances in Neural Information Processing Systems}, 36, 2024.

\bibitem[Nagel et~al.(2020)Nagel, Amjad, Van~Baalen, Louizos, and Blankevoort]{nagel2020up}
Markus Nagel, Rana~Ali Amjad, Mart Van~Baalen, Christos Louizos, and Tijmen Blankevoort.
\newblock Up or down? adaptive rounding for post-training quantization.
\newblock In \emph{International Conference on Machine Learning}, pages 7197--7206. PMLR, 2020.

\bibitem[Netzer et~al.(2011)Netzer, Wang, Coates, Bissacco, Wu, Ng, et~al.]{netzer2011reading}
Yuval Netzer, Tao Wang, Adam Coates, Alessandro Bissacco, Baolin Wu, Andrew~Y Ng, et~al.
\newblock Reading digits in natural images with unsupervised feature learning.
\newblock In \emph{NIPS workshop on deep learning and unsupervised feature learning}, page~4. Granada, 2011.

\bibitem[Nguyen et~al.(2022)Nguyen, Huynh, Nguyen, Liew, Yin, and Nguyen]{nguyen2022survey}
Thanh~Tam Nguyen, Thanh~Trung Huynh, Phi~Le Nguyen, Alan Wee-Chung Liew, Hongzhi Yin, and Quoc Viet~Hung Nguyen.
\newblock A survey of machine unlearning.
\newblock \emph{arXiv preprint arXiv:2209.02299}, 2022.

\bibitem[Sekhari et~al.(2021)Sekhari, Acharya, Kamath, and Suresh]{sekhari2021remember}
Ayush Sekhari, Jayadev Acharya, Gautam Kamath, and Ananda~Theertha Suresh.
\newblock Remember what you want to forget: Algorithms for machine unlearning.
\newblock \emph{Advances in Neural Information Processing Systems}, 34:\penalty0 18075--18086, 2021.

\bibitem[Shokri et~al.(2017)Shokri, Stronati, Song, and Shmatikov]{shokri2017membership}
Reza Shokri, Marco Stronati, Congzheng Song, and Vitaly Shmatikov.
\newblock Membership inference attacks against machine learning models.
\newblock In \emph{2017 IEEE symposium on security and privacy (SP)}, pages 3--18. IEEE, 2017.

\bibitem[Thudi et~al.(2022)Thudi, Deza, Chandrasekaran, and Papernot]{thudi2022unrolling}
Anvith Thudi, Gabriel Deza, Varun Chandrasekaran, and Nicolas Papernot.
\newblock Unrolling sgd: Understanding factors influencing machine unlearning.
\newblock In \emph{2022 IEEE 7th European Symposium on Security and Privacy (EuroS\&P)}, pages 303--319. IEEE, 2022.

\bibitem[Vaswani et~al.(2017)Vaswani, Shazeer, Parmar, Uszkoreit, Jones, Gomez, Kaiser, and Polosukhin]{vaswani2017attention}
Ashish Vaswani, Noam Shazeer, Niki Parmar, Jakob Uszkoreit, Llion Jones, Aidan~N Gomez, {\L}ukasz Kaiser, and Illia Polosukhin.
\newblock Attention is all you need.
\newblock \emph{Advances in neural information processing systems}, 30, 2017.

\bibitem[Warnecke et~al.(2021)Warnecke, Pirch, Wressnegger, and Rieck]{warnecke2021machine}
Alexander Warnecke, Lukas Pirch, Christian Wressnegger, and Konrad Rieck.
\newblock Machine unlearning of features and labels.
\newblock \emph{arXiv preprint arXiv:2108.11577}, 2021.

\bibitem[Xu et~al.(2020)Xu, Li, Zhuang, Liu, Cao, Liang, and Tan]{xu2020generative}
Shoukai Xu, Haokun Li, Bohan Zhuang, Jing Liu, Jiezhang Cao, Chuangrun Liang, and Mingkui Tan.
\newblock Generative low-bitwidth data free quantization.
\newblock In \emph{Computer Vision--ECCV 2020: 16th European Conference, Glasgow, UK, August 23--28, 2020, Proceedings, Part XII 16}, pages 1--17. Springer, 2020.

\bibitem[Zhang et~al.(2021)Zhang, Qin, Ding, Gong, Yan, Tao, Li, Yu, and Liu]{zhang2021diversifying}
Xiangguo Zhang, Haotong Qin, Yifu Ding, Ruihao Gong, Qinghua Yan, Renshuai Tao, Yuhang Li, Fengwei Yu, and Xianglong Liu.
\newblock Diversifying sample generation for accurate data-free quantization.
\newblock In \emph{Proceedings of the IEEE/CVF conference on computer vision and pattern recognition}, pages 15658--15667, 2021.

\bibitem[Zhou et~al.(2018)Zhou, Moosavi-Dezfooli, Cheung, and Frossard]{zhou2018adaptive}
Yiren Zhou, Seyed-Mohsen Moosavi-Dezfooli, Ngai-Man Cheung, and Pascal Frossard.
\newblock Adaptive quantization for deep neural network.
\newblock In \emph{Proceedings of the AAAI Conference on Artificial Intelligence}, 2018.

\end{thebibliography}
}

\clearpage
\setcounter{page}{1}
\maketitlesupplementary

\begingroup
\renewcommand{\thesection}{\Alph{section}}
\setcounter{section}{0} 

\noindent The organization of the appendix is as follows:
\begin{itemize}[leftmargin=*]
    \item Appendix A: Additional  Implementation Details;
    \item Appendix B: ResNet18 on Tiny-Imagenet dataset;
    \item Appendix C: ResNet18 under Class-wise Forgetting;
    \item Appendix D: MobileNetV2 on CIFAR-10, CIFAR-100, SVHN datasets;
    \item Appendix E: Efficiency Analysis.
\end{itemize}

\section{Additional  Implementation Details} We adopte the experimental settings of SalUn and $\ell_1$-sparse   for the baseline methods. All experiments use the SGD optimizer. For FT and RL, we train for 10 epochs within the interval [1e-3, 1e-1]. For GA, we train for 5 epochs with learning rate within the interval [1e-5, 1e-3]. For IU, we explore the parameter $\alpha$
associated with the woodfisher Hessian Inverse approximation within the range [1, 20]. For $\ell_1$-sparse,
a learning rate search for the parameter $\gamma$ is executed within the range [1e-6, 1e-4], while searching
for the learning rate within the range [1e-3, 1e-1].  For SalUn , we train for 10 epochs with learning rates in the range [5e-4, 5e-2] and sparsity ratios in the range [0.1, 0.9]. For Q-MUL, we train the unlearned model using the SGD optimizer with a batch size of 256. In the random data forgetting scenario, for ResNet-18, we train for 10 epochs with learning rates in the range [1e-3, 1e-1]. For MobileNetV2, train for 10 epochs with learning rates in the range [1e-2, 1e-1]. In the classwise forgetting scenario, for ResNet-18, we train for 10 epochs with learning rates in the range [1e-3, 1e-1]. All experiments are conducted on a single NVIDIA RTX 4090 GPU.

\section{ResNet18 on Tiny-Imagenet dataset}
Table \ref{tab4} shows the experimental results of the quantized ResNet18 on the larger dataset Tiny-ImageNet. On the larger dataset, Q-MUL can also demonstrate superior performance. Specifically, the gaps between Q-MUL and Retrain in the four metrics FA, RA, TA, and MIA are 9.95\%, 7.47\%, 6.32\%, and 8.95\%, respectively. Compared with other methods, the average gap is as low as 8.17\%. This indicates that Q-MUL can also exhibit excellent performance on the larger datasets.

\begin{figure}[t]
\centering

\includegraphics[width=0.9\columnwidth]{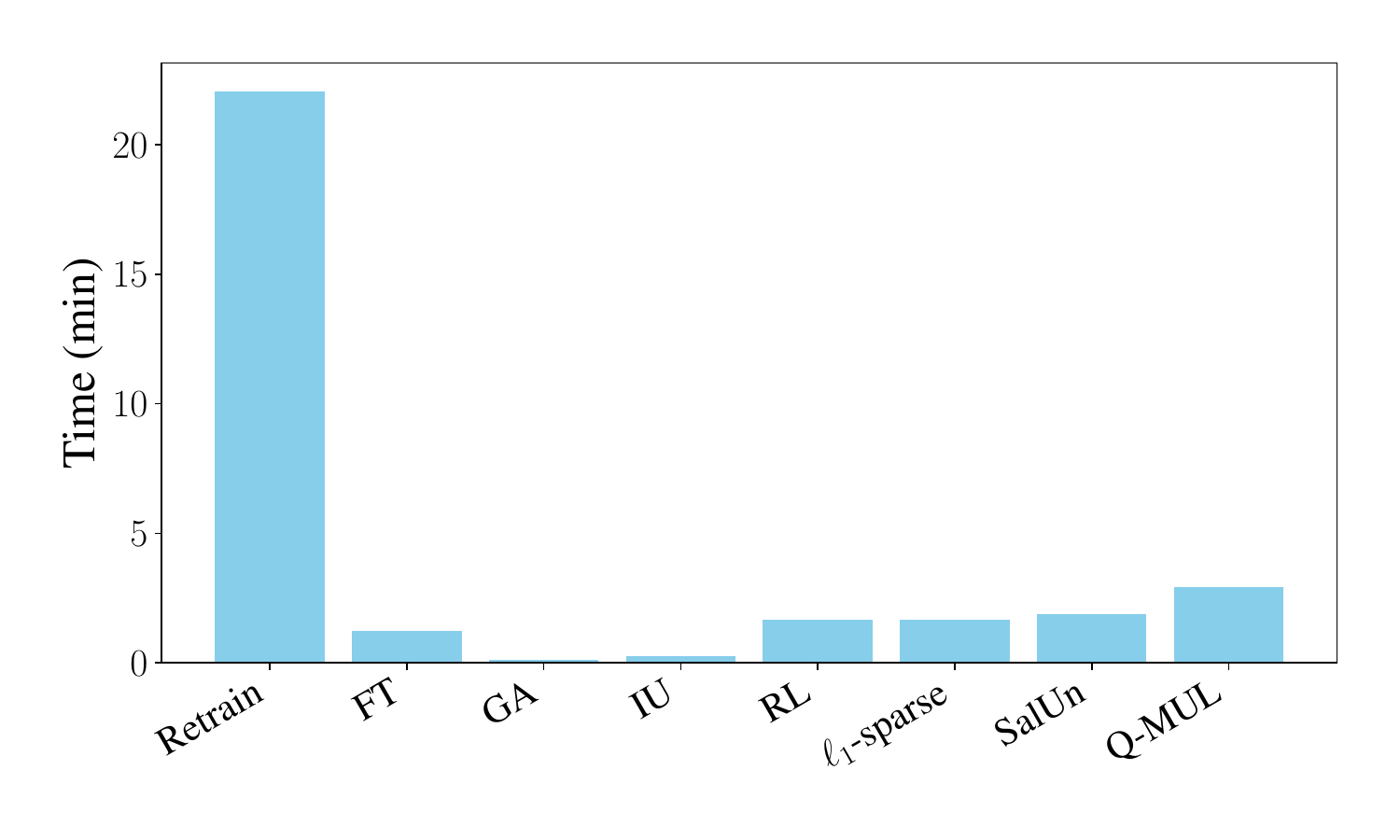} 
\caption{ 
Efficiency Analysis. The experimental scenario involves MobileNetV2 performing random data forgetting (10\%) on CIFAR-100. 
}
\label{fig4}
\end{figure}

\section{ResNet18 under Class-wise Forgetting}
Tables \ref{tab5} and \ref{tab6} show the experiments of the quantized ResNet18 under Class-wise forgetting scenario. We conduct experiments on two datasets: CIFAR-10 and CIFAR-100. For the class-wise forgetting scenario, RL, $\ell_1$-sparse, SalUn, and Q-MUL all perform very close to Retrain. Specifically, on the CIFAR-10, the average gaps of these methods from Retrain are 0.13\%, 0.08\%, 0.10\%, and 0.11\% respectively. On the CIFAR-100, the average gaps of these methods from Retrain are 0.76\%, 0.55\%, 0.97\%, and 0.32\%, respectively. 

\section{MobileNetV2 on CIFAR-10, CIFAR-100, SVHN datasets}
We present the experiments of the quantized MobileNetV2 on CIFAR-10, CIFAR-100, and SVHN in Tables \ref{tab7}, \ref{tab8}, and \ref{tab9}. For MobileNetV2 on CIFAR-10, Q-MUL achieves the best or second-best performance.  For MobileNetV2 on CIFAR-100, the average gaps of Q-MUL from Retrain at data forgetting ratios of 10\%, 30\%, and 50\% are 6.52\%, 9.07\%, and 10.53\%, respectively. Compared to the previous state-of-the-art method, these gaps are reduced by 2.77\%, 3.19\%, and 0.70\% respectively. For MobileNetV2 on SVHN, at data forgetting ratios of 10\%, 30\%, and 50\%, the average gaps of Q-MUL from Retrain are 0.92\%, 1.73\%, and 3.49\% respectively.

\begin{table*}[h!]
  \centering
    \begin{tabular}{c|ccccc}
\toprule   
\multirow{2}{*}{Method}  & \multicolumn{5}{c}{Tiny-Imagenet}  \\
\cmidrule{2-6}
  & FA & RA & TA & MIA & AG$\color{blue}\downarrow$ \\
\midrule
    Retrain &62.15&99.46&62.65&51.51&\cellcolor{gray!20}0 \\
   FT &74.45\textcolor{blue}{(12.30)}&91.99\textcolor{blue}{(7.47)}&56.57\textcolor{blue}{(6.08)}&33.67\textcolor{blue}{(17.84)}&10.92 \cellcolor{gray!20}\\
    GA &92.95\textcolor{blue}{(30.8)}&93.01\textcolor{blue}{(6.45)}&57.19\textcolor{blue}{(5.46)}&11.77\textcolor{blue}{(39.74)}&20.61\cellcolor{gray!20}  \\
    IU &0.45\textcolor{blue}{(61.70)}&0.51\textcolor{blue}{(98.95)}&0.50\textcolor{blue}{(62.15)}&0.45\textcolor{blue}{(51.06)}&68.47\cellcolor{gray!20} \\
    RL &60.05\textcolor{blue}{(2.10)}&82.60\textcolor{blue}{(16.86)}&51.59\textcolor{blue}{(11.06)}&45.13\textcolor{blue}{(6.38)}&9.10\cellcolor{gray!20} \\
    $\ell_1$-sparse &60.89\textcolor{blue}{(1.26)}&75.44\textcolor{blue}{(24.02)}&56.41\textcolor{blue}{(6.24)}&46.94\textcolor{blue}{(4.57)}&\underline{9.02}\cellcolor{gray!20} \\
    SalUn &65.28\textcolor{blue}{(3.13)}&81.70\textcolor{blue}{(17.76)}&51.77\textcolor{blue}{(10.88)}&37.08\textcolor{blue}{(14.43)}&11.55\cellcolor{gray!20} \\
    Q-MUL &72.10\textcolor{blue}{(9.95)}&91.99\textcolor{blue}{(7.47)}&56.53\textcolor{blue}{(6.32)}&42.56\textcolor{blue}{(8.95)}&\textbf{8.17}\cellcolor{gray!20} \\
    \bottomrule
    \end{tabular}%
     \caption{Performance of various MU methods for ResNet18 with 4-bit quantization on Tiny-Imagenet. The unlearning scenario is random data forgetting(10\%).  \textbf{Bold} indicates the best performance and \underline{underline} indicates the runner-up.}
  \label{tab4}
\end{table*}

\begin{table*}[h!]
  \centering
    \begin{tabular}{c|ccccc}
\toprule
\multirow{2}{*}{Method}  & \multicolumn{5}{c}{CIFAR-10}  \\
\cmidrule{2-6}
  & FA & RA & TA & MIA & AG$\color{blue}\downarrow$ \\
\midrule
Retrain  & 0.00 & 99.99 & 93.34 & 100.0 &\cellcolor{gray!20}0 \\
FT       & 62.40\textcolor{blue}{(62.40)} & 99.95\textcolor{blue}{(0.04)} & 93.59\textcolor{blue}{(0.25)} & 94.67\textcolor{blue}{(5.33)} &17.02\cellcolor{gray!20} \\
GA       & 11.20\textcolor{blue}{(11.20)} & 92.28\textcolor{blue}{(7.71)} & 85.49\textcolor{blue}{(7.85)} & 92.13\textcolor{blue}{(7.87)} &8.66\cellcolor{gray!20} \\
IU       & 0.00\textcolor{blue}{(0.00)} & 29.75\textcolor{blue}{(70.24)} & 28.92\textcolor{blue}{(64.42)} & 100.0\textcolor{blue}{(0.00)} &33.67\cellcolor{gray!20} \\
RL       & 0.00\textcolor{blue}{(0.00)} & 99.87\textcolor{blue}{(0.12)} & 93.74\textcolor{blue}{(0.40)} & 100.0\textcolor{blue}{(0.00)} &0.13\cellcolor{gray!20} \\
$\ell_1$-sparse & 0.00\textcolor{blue}{(0.00)} & 99.85\textcolor{blue}{(0.14)} & 93.51\textcolor{blue}{(0.17)} & 100.0\textcolor{blue}{(0.00)} &\textbf{0.08}\cellcolor{gray!20} \\
SalUn    & 0.04\textcolor{blue}{(0.04)} & 99.84\textcolor{blue}{(0.15)} & 93.56\textcolor{blue}{(0.22)} & 100.0\textcolor{blue}{(0.00)} &\underline{0.10}\cellcolor{gray!20} \\
Q-MUL    & 0.00\textcolor{blue}{(0.00)} & 99.69\textcolor{blue}{(0.30)} & 93.21\textcolor{blue}{(0.13)} & 100.0\textcolor{blue}{(0.00)} &0.11\cellcolor{gray!20} \\
\bottomrule
\end{tabular}%
\caption{Performance of various MU methods for ResNet18 with 4-bit quantization on CIFAR-10. The unlearning scenario is class-wise forgetting (one class is forgotten.). \textbf{Bold} indicates the best performance and \underline{underline} indicates the runner-up.}
\label{tab5}
\end{table*}

\begin{table*}[h!]
  \centering
    \begin{tabular}{c|ccccc}
\toprule
\multirow{2}{*}{Method}  & \multicolumn{5}{c}{CIFAR-100}  \\
\cmidrule{2-6}
  & FA & RA & TA & MIA & AG$\color{blue}\downarrow$ \\
\midrule
Retrain  & 0.00 & 99.96 & 70.63 & 100.0 &\cellcolor{gray!20}0 \\
FT       & 7.11\textcolor{blue}{(7.11)} & 99.79\textcolor{blue}{(0.27)} & 71.10\textcolor{blue}{(0.47)} & 100.0\textcolor{blue}{(0.00)} &1.96\cellcolor{gray!20} \\
GA       & 37.56\textcolor{blue}{(37.56)} & 88.69\textcolor{blue}{(11.27)} & 61.75\textcolor{blue}{(8.88)} & 79.78\textcolor{blue}{(20.22)} &19.48\cellcolor{gray!20} \\
IU       & 0.00\textcolor{blue}{(0.00)} & 72.25\textcolor{blue}{(17.71)} & 51.22\textcolor{blue}{(19.41)} & 100.0\textcolor{blue}{(0.00)} &9.28\cellcolor{gray!20} \\
RL       & 0.89\textcolor{blue}{(0.89)} & 99.97\textcolor{blue}{(0.01)} & 72.78\textcolor{blue}{(2.15)} & 100.0\textcolor{blue}{(0.00)} &0.76\cellcolor{gray!20} \\
$\ell_1$-sparse & 0.00\textcolor{blue}{(0.00)} & 98.33\textcolor{blue}{(1.63)} & 70.06\textcolor{blue}{(0.57)} & 100.0\textcolor{blue}{(0.00)} &\underline{0.55}\cellcolor{gray!20} \\
SalUn    & 1.33\textcolor{blue}{(1.33)} & 99.92\textcolor{blue}{(0.04)} & 73.12\textcolor{blue}{(2.49)} & 100.0\textcolor{blue}{(0.00)} &0.97\cellcolor{gray!20} \\
Q-MUL    & 0.00\textcolor{blue}{(0.00)} & 99.50\textcolor{blue}{(0.46)} & 69.79\textcolor{blue}{(0.84)} & 100.0\textcolor{blue}{(0.00)} &\textbf{0.32}\cellcolor{gray!20} \\
\bottomrule
\end{tabular}%
\caption{Performance of various MU methods for ResNet18 with 4-bit quantization on CIFAR-100. The unlearning scenario is class-wise forgetting (one class is forgotten). \textbf{Bold} indicates the best performance and \underline{underline} indicates the runner-up.}
\label{tab6}
\end{table*}

\section{Efficiency Analysis}
We conduct an efficiency analysis of various unlearning methods. As shown in Figure  \ref{fig4}, Retrain consumes a significant amount of time, requiring 22.05 minutes. Although GA and IU have minimal time overhead, these two methods perform extremely poorly on quantized models during unlearning, making them almost unusable. Q-MUL, due to the need to calculate the gradients of the forgotten dataset and the retained dataset on the loss, has a slightly larger time overhead compared to methods like SalUn and RL, but the unlearning effect is significantly improved. Q-MUL achieves a trade-off between unlearning effectiveness and time overhead.

\begin{table*}[b]
  \centering
    \begin{tabular}{c|ccccc}
\toprule  
\multirow{2}[4]{*}{Method}  & \multicolumn{5}{c}{CIFAR-10} \\
\cmidrule{2-6}
   & FA & RA & TA & MIA & AG$\color{blue}\downarrow$ \\
    \midrule
    \multicolumn{6}{c}{\cellcolor{gray!20}The proportion of forgotten data samples to all samples is 10\%}\\
    \midrule
    Retrain  & 85.97 & 94.39 & 85.49 & 18.60 &\cellcolor{gray!20} 0  \\
    FT  & 93.64 \textcolor{blue}{(7.67)} & 95.42 \textcolor{blue}{(1.03)} & 87.37 \textcolor{blue}{(1.88)} & 9.62 \textcolor{blue}{(8.98)} &\cellcolor{gray!20} 4.89   \\
   GA  & 93.97 \textcolor{blue}{(8.00)} & 94.87 \textcolor{blue}{(0.48)} & 87.21 \textcolor{blue}{(1.72)} & 8.49 \textcolor{blue}{(10.11)} &\cellcolor{gray!20} 5.08  \\
    IU &  13.13\textcolor{blue}{(72.84)} & 14.28\textcolor{blue}{(80.11)}  & 13.97\textcolor{blue}{(71.52)}  & 84.51\textcolor{blue}{(66.91)}  &\cellcolor{gray!20} 75.85  \\
    RL &  85.82 \textcolor{blue}{(0.15)}& 87.46 \textcolor{blue}{(6.93)}& 84.05 \textcolor{blue}{(1.44)}& 17.27 \textcolor{blue}{(1.33)}&\cellcolor{gray!20} \textbf{2.46} \\
    $\ell_1$-sparse & 93.17 \textcolor{blue}{(7.20)} & 95.15 \textcolor{blue}{(0.76)} & 87.66 \textcolor{blue}{(2.17)} & 9.89 \textcolor{blue}{(8.71)} &\cellcolor{gray!20} 4.71 \\
    SalUn  & 90.87\textcolor{blue}{(4.90)} & 92.08\textcolor{blue}{(2.31)} & 86.63\textcolor{blue}{(1.14)} & 15.71\textcolor{blue}{(2.89)} &\cellcolor{gray!20} 2.81 \\
    Q-MUL  & 90.71\textcolor{blue}{(4.74)} & 93.60\textcolor{blue}{(0.79)} & 87.51\textcolor{blue}{(2.02)} & 15.00\textcolor{blue}{(3.60)} &\cellcolor{gray!20} \underline{2.79} \\
    \midrule
    \multirow{2}[4]{*}{Method}  & \multicolumn{5}{c}{CIFAR-10} \\
\cmidrule{2-6}
   & FA & RA & TA & MIA & AG$\color{blue}\downarrow$ \\
    \midrule
    \multicolumn{6}{c}{\cellcolor{gray!20}The proportion of forgotten data samples to all samples is 30\%}\\
    \midrule
    Retrain  & 84.87 & 93.87 & 84.19 & 19.90 &\cellcolor{gray!20} 0   \\
    FT  & 91.53 \textcolor{blue}{(6.66)} & 93.17 \textcolor{blue}{(0.70)} & 86.21 \textcolor{blue}{(2.02)} & 12.07 \textcolor{blue}{(7.83)} &\cellcolor{gray!20} 4.30 \\
   GA  & 89.16 \textcolor{blue}{(4.29)} & 89.40 \textcolor{blue}{(4.47)} & 82.79 \textcolor{blue}{(1.40)} & 15.20 \textcolor{blue}{(4.70)} &\cellcolor{gray!20} 3.72 \\
    IU  & 12.67\textcolor{blue}{(72.20)}  & 13.11\textcolor{blue}{(80.76)}  & 12.65\textcolor{blue}{(71.54)}  & 86.47\textcolor{blue}{(66.57)}  &\cellcolor{gray!20} 72.77 \\
    RL  & 88.04 \textcolor{blue}{(3.17)}& 88.86 \textcolor{blue}{(5.01)}& 84.26 \textcolor{blue}{(0.07)}& 21.81\textcolor{blue}{(1.91)}  &\cellcolor{gray!20} \underline{2.54} \\
    $\ell_1$-sparse & 92.90 \textcolor{blue}{(8.03)} & 94.53 \textcolor{blue}{(0.66)} & 86.80 \textcolor{blue}{(2.61)} & 10.76 \textcolor{blue}{(9.14)} &\cellcolor{gray!20} 5.11 \\
    SalUn  & 95.20\textcolor{blue}{(10.33)} & 96.83\textcolor{blue}{(2.96)} & 90.49\textcolor{blue}{(6.30)} & 16.87\textcolor{blue}{(3.03)} &\cellcolor{gray!20} 5.66 \\
    Q-MUL  & 87.56\textcolor{blue}{(2.69)} & 89.24\textcolor{blue}{(4.63)} & 85.63\textcolor{blue}{(1.44)} & 18.56\textcolor{blue}{(1.34)} &\cellcolor{gray!20} \textbf{2.53} \\
    \midrule
    \multirow{2}[4]{*}{Method}  & \multicolumn{5}{c}{CIFAR-10} \\
\cmidrule{2-6}
   & FA & RA & TA & MIA & AG$\color{blue}\downarrow$ \\
    \midrule
    \multicolumn{6}{c}{\cellcolor{gray!20}The proportion of forgotten data samples to all samples is 50\%}\\
    \midrule
    Retrain  & 82.27 & 94.76 & 82.17 & 22.17 & \cellcolor{gray!20} 0   \\
    FT  & 89.50 \textcolor{blue}{(7.23)} & 91.26 \textcolor{blue}{(3.50)} & 84.32 \textcolor{blue}{(2.15)} & 13.05 \textcolor{blue}{(9.12)} & \cellcolor{gray!20}5.5 \\
   GA  & 75.99 \textcolor{blue}{(6.28)} & 76.07 \textcolor{blue}{(18.69)} & 72.07 \textcolor{blue}{(10.10)} & 24.30 \textcolor{blue}{(2.13)} &\cellcolor{gray!20} 9.3 \\
    IU  & 21.89\textcolor{blue}{(60.38)} & 22.03\textcolor{blue}{(72.73)} & 21.67\textcolor{blue}{(60.50)} & 78.52\textcolor{blue}{(56.35)} &\cellcolor{gray!20} 62.49 \\
    RL & 86.81 \textcolor{blue}{(4.54)} & 87.85 \textcolor{blue}{(6.91)}& 83.70\textcolor{blue}{(1.53)} & 17.98 \textcolor{blue}{(4.19)}&\cellcolor{gray!20} \underline{4.29} \\
    $\ell_1$-sparse & 6.64 \textcolor{blue}{(11.09)} & 95.70 \textcolor{blue}{(0.94)} & 87.11 \textcolor{blue}{(4.94)} & 10.00 \textcolor{blue}{(12.17)} &\cellcolor{gray!20} 7.29  \\
    SalUn  & 87.98\textcolor{blue}{(5.71)} & 88.95\textcolor{blue}{(5.81)} & 84.56\textcolor{blue}{(2.39)} & 15.04\textcolor{blue}{(7.13)} &\cellcolor{gray!20} 5.26 \\
    Q-MUL  & 87.08\textcolor{blue}{(4.81)} & 89.48\textcolor{blue}{(5.28)} & 84.76\textcolor{blue}{(2.59)} & 22.24\textcolor{blue}{(0.07)} &\cellcolor{gray!20} \textbf{3.19} \\
    \bottomrule
    \end{tabular}%
    \caption{Performance of various MU methods for MobileNetV2 with activations kept at full precision and weights quantized to 2 bits on
CIFAR-10. The unlearning scenario is random data forgetting. \textbf{Bold} indicates the best performance and \underline{underline} indicates the runner-up. A performance gap against Retrain is provided in \textcolor{blue}{(•)}.}
  \label{tab7}%
\end{table*}

\begin{table*}[b]
  \centering
    \begin{tabular}{c|ccccc}
\toprule  
\multirow{2}[4]{*}{Method}  & \multicolumn{5}{c}{CIFAR-100} \\
\cmidrule{2-6}
   & FA & RA & TA & MIA & AG$\color{blue}\downarrow$ \\
    \midrule
    \multicolumn{6}{c}{\cellcolor{gray!20}The proportion of forgotten data samples to all samples is 10\%}\\
    \midrule
    Retrain  & 61.53 & 89.44 & 60.90 & 40.20 &\cellcolor{gray!20} 0  \\
    FT  & 82.89 \textcolor{blue}{(21.36)} & 88.18 \textcolor{blue}{(1.26)} & 63.56 \textcolor{blue}{(2.66)} & 19.11 \textcolor{blue}{(21.09)} &\cellcolor{gray!20} 11.59\\
   GA  & 83.82 \textcolor{blue}{(22.29)} & 85.20 \textcolor{blue}{(4.24)} & 61.80 \textcolor{blue}{(0.90)} & 17.33 \textcolor{blue}{(22.87)} &\cellcolor{gray!20} 12.58\\
    IU &  2.40\textcolor{blue}{(59.13)}  & 2.81\textcolor{blue}{(86.63)}  & 2.76\textcolor{blue}{(58.14)}  & 2.62\textcolor{blue}{(37.58)}  &\cellcolor{gray!20} 60.37 \\
    RL &  81.53 \textcolor{blue}{(20.00)}&  88.41 \textcolor{blue}{(1.03)} & 63.26 \textcolor{blue}{(2.36)} & 26.38 \textcolor{blue}{(13.82)} &\cellcolor{gray!20} 9.30\\
    $\ell_1$-sparse & 81.31 \textcolor{blue}{(19.78)} & 85.97 \textcolor{blue}{(3.47)} & 62.33 \textcolor{blue}{(1.43)} & 19.98 \textcolor{blue}{(20.22)} &\cellcolor{gray!20} 11.23 \\
    SalUn  & 82.22\textcolor{blue}{(20.69)} & 88.15\textcolor{blue}{(1.29)} & 63.26\textcolor{blue}{(2.36)} &  27.38\textcolor{blue}{(12.82)} &\cellcolor{gray!20}\underline{9.29} \\
    Q-MUL  &69.96\textcolor{blue}{(8.43)}  &80.25\textcolor{blue}{(9.19)}  &62.29\textcolor{blue}{(1.39)}  &33.15\textcolor{blue}{(7.05)}  &\textbf{6.52}\cellcolor{gray!20}   \\
    \midrule
    \multirow{2}[4]{*}{Method}  & \multicolumn{5}{c}{CIFAR-100} \\
\cmidrule{2-6}
   & FA & RA & TA & MIA & AG$\color{blue}\downarrow$ \\
    \midrule
    \multicolumn{6}{c}{\cellcolor{gray!20}The proportion of forgotten data samples to all samples is 30\%}\\
    \midrule
    Retrain  & 53.90 & 83.38 & 54.97 & 46.46 &\cellcolor{gray!20} 0 \\
    FT  & 79.36 \textcolor{blue}{(25.46)} & 86.34 \textcolor{blue}{(2.96)} & 62.04 \textcolor{blue}{(7.07)} & 22.03 \textcolor{blue}{(24.43)} &\cellcolor{gray!20} 14.98\\
   GA  & 66.02 \textcolor{blue}{(12.12)} & 66.34 \textcolor{blue}{(17.04)} & 51.44 \textcolor{blue}{(3.53)} & 24.67 \textcolor{blue}{(21.79)} &\cellcolor{gray!20} 13.62 \\
    IU  & 6.68\textcolor{blue}{(47.22)}  & 7.30\textcolor{blue}{(76.08)}  & 6.75\textcolor{blue}{(48.22)}  & 94.62\textcolor{blue}{(48.16)}  &\cellcolor{gray!20} 54.92 \\
    RL  & 77.16 \textcolor{blue}{(23.26)}& 83.49\textcolor{blue}{(0.11)} & 61.31 \textcolor{blue}{(6.34)} & 25.43\textcolor{blue}{(21.03)}  &\cellcolor{gray!20} 12.69 \\
    $\ell_1$-sparse & 82.53 \textcolor{blue}{(28.63)} & 88.77 \textcolor{blue}{(5.39)} & 63.30 \textcolor{blue}{(8.33)} & 19.75 \textcolor{blue}{(26.71)} &\cellcolor{gray!20} 17.27 \\
    SalUn  & 76.26\textcolor{blue}{(22.36)} & 83.46\textcolor{blue}{(0.08)} & 61.19\textcolor{blue}{(6.22)} & 26.08\textcolor{blue}{(20.38)} &\cellcolor{gray!20}  \underline{12.26} \\
    Q-MUL  &67.77\textcolor{blue}{(13.87)}  &79.60\textcolor{blue}{(3.78)}  &61.99\textcolor{blue}{(7.02)}  &34.87\textcolor{blue}{(11.59)}  &\textbf{9.07}\cellcolor{gray!20}   \\
     \midrule
    \multirow{2}[4]{*}{Method}  & \multicolumn{5}{c}{CIFAR-100} \\
\cmidrule{2-6}
   & FA & RA & TA & MIA & AG$\color{blue}\downarrow$ \\
    \midrule
    \multicolumn{6}{c}{\cellcolor{gray!20}The proportion of forgotten data samples to all samples is 50\%}\\
    \midrule
    Retrain  & 49.20 & 85.03 & 50.21 & 51.44 &\cellcolor{gray!20} 0 \\
    FT  & 83.76 \textcolor{blue}{(34.56)} & 91.24 \textcolor{blue}{(6.21)}& 63.92 \textcolor{blue}{(13.71)}& 20.10\textcolor{blue}{(31.34)} &\cellcolor{gray!20} 18.65\\
   GA  & 41.20 \textcolor{blue}{(8.00)} & 40.60 \textcolor{blue}{(44.43)}& 33.94 \textcolor{blue}{(16.27)}& 46.70 \textcolor{blue}{(5.44)}&\cellcolor{gray!20} 18.54\\
    IU  & 1.40\textcolor{blue}{(47.80)}  & 1.51\textcolor{blue}{(83.52)}  & 1.26\textcolor{blue}{(48.95)}  & 22\textcolor{blue}{(29.44)}  &\cellcolor{gray!20} 52.43 \\
    RL & 75.36 \textcolor{blue}{(26.16)} & 80.92\textcolor{blue}{(4.11)} & 60.57\textcolor{blue}{(10.36)} & 33.59\textcolor{blue}{(17.85)} &\cellcolor{gray!20}  14.62 \\
    $\ell_1$-sparse & 18.69 \textcolor{blue}{(19.78)} & 85.97 \textcolor{blue}{(3.47)} & 62.33 \textcolor{blue}{(1.43)} & 19.98 \textcolor{blue}{(20.22)} &\cellcolor{gray!20} \underline{11.23} \\
    SalUn  & 73.56\textcolor{blue}{(24.36)} & 79.50\textcolor{blue}{(5.53)} & 59.31\textcolor{blue}{(9.10)} & 34.78\textcolor{blue}{(16.66)} &\cellcolor{gray!20} 13.91 \\
    Q-MUL  &67.93\textcolor{blue}{(18.73)}  &78.83\textcolor{blue}{(6.20)}  &60.38\textcolor{blue}{(10.17)}  &44.41\textcolor{blue}{(7.03)}  &\textbf{10.53}\cellcolor{gray!20}    \\
    \bottomrule
    \end{tabular}%
    \caption{Performance of various MU methods for MobileNetV2 with activations kept at full precision and weights quantized to 2 bits on
CIFAR-100. The unlearning scenario is random data forgetting. \textbf{Bold} indicates the best performance and \underline{underline} indicates the runner-up. A performance gap against Retrain is provided in \textcolor{blue}{(•)}.}
  \label{tab8}%
\end{table*}

\begin{table*}[htbp]
  \centering
  \begin{tabular}{c|ccccc}
  \toprule
    \multirow{2}[4]{*}{Method}  & \multicolumn{5}{c}{SVHN} \\
\cmidrule{2-6}
   & FA & RA & TA & MIA & AG$\color{blue}\downarrow$ \\
  \midrule
  \multicolumn{6}{c}{\cellcolor{gray!20}The proportion of forgotten data samples to all samples is 10\%}\\
  \midrule
  Retrain & 94.25 & 99.99 & 94.17 & 9.33 & 0 \cellcolor{gray!20}\\
  FT      & 99.12 \textcolor{blue}{(4.87)} & 99.98 \textcolor{blue}{(0.01)} & 94.71 \textcolor{blue}{(0.54)} & 2.18 \textcolor{blue}{(7.15)} & 3.14 \cellcolor{gray!20}\\
  GA      & 99.21 \textcolor{blue}{(4.96)} & 99.43 \textcolor{blue}{(0.56)} & 94.84 \textcolor{blue}{(0.67)} & 10.77 \textcolor{blue}{(1.44)} & \underline{1.91}\cellcolor{gray!20} \\
  IU      & 98.65 \textcolor{blue}{(4.40)} & 98.83 \textcolor{blue}{(1.16)} & 93.96 \textcolor{blue}{(0.21)} & 2.43 \textcolor{blue}{(6.90)} & 3.17\cellcolor{gray!20} \\
  RL      & 96.81 \textcolor{blue}{(2.56)} & 98.36 \textcolor{blue}{(1.63)} & 94.46 \textcolor{blue}{(0.29)} & 20.20 \textcolor{blue}{(10.87)} & 3.84\cellcolor{gray!20} \\
  $\ell_1$-sparse & 99.07 \textcolor{blue}{(4.82)} & 99.97 \textcolor{blue}{(0.02)} & 94.67 \textcolor{blue}{(0.50)} & 2.68 \textcolor{blue}{(6.65)} & 3.00\cellcolor{gray!20} \\
  SalUn   & 96.42 \textcolor{blue}{(2.17)} & 98.19 \textcolor{blue}{(1.80)} & 94.51 \textcolor{blue}{(0.34)} & 20.39 \textcolor{blue}{(11.06)} & 3.84\cellcolor{gray!20} \\
  Q-MUL  & 96.47 \textcolor{blue}{(2.22)} & 99.48 \textcolor{blue}{(0.51)} & 94.96 \textcolor{blue}{(0.79)} & 9.16 \textcolor{blue}{(0.17)} & \textbf{0.92} \cellcolor{gray!20}\\
  \midrule
   \multirow{2}[4]{*}{Method}  & \multicolumn{5}{c}{SVHN} \\
\cmidrule{2-6}
   & FA & RA & TA & MIA & AG$\color{blue}\downarrow$ \\
  \midrule
  \multicolumn{6}{c}{\cellcolor{gray!20}The proportion of forgotten data samples to all samples is 30\%}\\
  \midrule
  Retrain & 93.12 & 100.0 & 93.81 & 11.21 & 0\cellcolor{gray!20} \\
  FT      & 99.28 \textcolor{blue}{(6.16)} & 99.98 \textcolor{blue}{(0.02)} & 94.72 \textcolor{blue}{(0.91)} & 2.13 \textcolor{blue}{(9.08)} & 4.04 \cellcolor{gray!20}\\
  GA      & 99.27 \textcolor{blue}{(6.15)} & 99.46 \textcolor{blue}{(0.54)} & 94.77 \textcolor{blue}{(0.96)} & 1.08 \textcolor{blue}{(10.13)} & 4.45 \cellcolor{gray!20}\\
  IU      & 98.63 \textcolor{blue}{(5.41)} & 98.83 \textcolor{blue}{(1.17)} & 93.05 \textcolor{blue}{(0.76)} & 2.92 \textcolor{blue}{(8.29)} & 3.91 \cellcolor{gray!20}\\
  RL      & 94.55 \textcolor{blue}{(1.43)} & 96.99 \textcolor{blue}{(3.01)} & 93.79 \textcolor{blue}{(0.02)} & 22.33 \textcolor{blue}{(11.12)} & \underline{3.90} \cellcolor{gray!20}\\
  $\ell_1$-sparse & 99.21 \textcolor{blue}{(6.09)} & 99.96 \textcolor{blue}{(0.04)} & 94.81 \textcolor{blue}{(1.00)} & 2.53 \textcolor{blue}{(8.68)} & 3.95\cellcolor{gray!20} \\
  SalUn   & 95.33 \textcolor{blue}{(2.21)} & 96.75 \textcolor{blue}{(3.25)} & 93.88 \textcolor{blue}{(0.07)} & 25.44 \textcolor{blue}{(14.23)} & 4.94 \cellcolor{gray!20}\\
  Q-MUL  & 96.80 \textcolor{blue}{(3.68)} & 99.46 \textcolor{blue}{(0.54)} & 94.95 \textcolor{blue}{(1.14)} & 12.77 \textcolor{blue}{(1.56)} & \textbf{1.73} \cellcolor{gray!20}\\
  \midrule
   \multirow{2}[4]{*}{Method}  & \multicolumn{5}{c}{SVHN} \\
\cmidrule{2-6}
   & FA & RA & TA & MIA & AG$\color{blue}\downarrow$ \\
  \midrule
  \multicolumn{6}{c}{\cellcolor{gray!20}The proportion of forgotten data samples to all samples is 50\%}\\
  \midrule
  Retrain & 92.57 & 95.83 & 93.19 & 12.46 & 0 \cellcolor{gray!20}\\
  FT      & 99.26 \textcolor{blue}{(6.69)} & 99.98 \textcolor{blue}{(4.15)} & 94.59 \textcolor{blue}{(1.40)} & 2.03 \textcolor{blue}{(10.43)} & 5.67\cellcolor{gray!20} \\
  GA      & 99.31 \textcolor{blue}{(6.74)} & 99.48 \textcolor{blue}{(3.65)} & 94.75 \textcolor{blue}{(1.56)} & 10.53 \textcolor{blue}{(1.93)} & \underline{3.47}\cellcolor{gray!20} \\
  IU      & 97.82 \textcolor{blue}{(5.25)} & 97.87 \textcolor{blue}{(2.04)} & 92.90 \textcolor{blue}{(0.29)} & 3.92 \textcolor{blue}{(8.54)} & 4.03\cellcolor{gray!20} \\
  RL      & 94.23 \textcolor{blue}{(1.66)} & 95.36 \textcolor{blue}{(0.47)} & 93.05 \textcolor{blue}{(0.14)} & 32.43 \textcolor{blue}{(19.97)} & 5.56\cellcolor{gray!20} \\
  $\ell_1$-sparse & 99.27 \textcolor{blue}{(6.70)} & 99.97 \textcolor{blue}{(4.14)} & 94.57 \textcolor{blue}{(1.38)} & 2.44 \textcolor{blue}{(10.02)} & 5.56\cellcolor{gray!20} \\
  SalUn   & 93.93 \textcolor{blue}{(1.36)} & 95.02 \textcolor{blue}{(0.81)} & 92.87 \textcolor{blue}{(0.32)} & 33.78 \textcolor{blue}{(21.32)} & 5.95\cellcolor{gray!20} \\
  Q-MUL  & 95.30 \textcolor{blue}{(2.73)} & 98.50 \textcolor{blue}{(2.67)} & 94.31 \textcolor{blue}{(1.12)} & 19.90 \textcolor{blue}{(7.44)} & \textbf{3.49} \cellcolor{gray!20}\\
  \bottomrule
  \end{tabular}
   \caption{Performance of various MU methods for MobileNetV2 with activations kept at full precision and weights quantized to 2 bits on
SVHN. The unlearning scenario is random data forgetting. \textbf{Bold} indicates the best performance and \underline{underline} indicates the runner-up. A performance gap against Retrain is provided in \textcolor{blue}{(•)}.}
  \label{tab9}
\end{table*}


\endgroup

\end{document}